%% file: main.tex
\documentclass[10pt,twocolumn,letterpaper,dvipsnames,table]{article}

\usepackage{iccv}

\makeatletter
\@namedef{ver@everyshi.sty}{}
\makeatother
\usepackage{tikz}
\usetikzlibrary{shapes.geometric, decorations.pathreplacing, calc}

\usepackage{xcolor}
\usepackage{times}
\usepackage{epsfig}
\usepackage{graphicx}
\usepackage{amsmath, amsbsy}
\usepackage{amssymb}
\usepackage{siunitx}
\usepackage{afterpage}
\usepackage[utf8]{inputenc}
\usepackage{url}
\usepackage{array}
\usepackage{hhline}
\usepackage{multirow}
\usepackage{courier}
\usepackage{bbm}
\usepackage{enumitem}
\usepackage{algorithm}
\usepackage[noend]{algpseudocode}
\usepackage{ctable}
\usepackage{color}

\usepackage{booktabs}
\usepackage{multirow}

\usepackage{subcaption}
\usepackage[pagebackref=true,breaklinks=true,letterpaper=true,colorlinks,bookmarks=false]{hyperref}

\usepackage{cleveref}

\input{commands}

\iccvfinalcopy % *** Uncomment this line for the final submission

\def\iccvPaperID{7768} % *** Enter the ICCV Paper ID here

\ificcvfinal\fi

\begin{document}
%%%%%%%%% TITLE
\title{DiagViB-6: A Diagnostic Benchmark Suite for 
Vision Models in the Presence of
Shortcut and Generalization Opportunities}

\author{Elias Eulig\textsuperscript{1,2,}\thanks{Corresponding authors. Contact via {\tt\small elias@eeulig.com} or {\tt\small volker.fischer@de.bosch.com}.}\quad Piyapat Saranrittichai\textsuperscript{1,3}\quad Chaithanya Kumar Mummadi\textsuperscript{1,3}\\
Kilian Rambach\textsuperscript{1}\quad William Beluch\textsuperscript{1} \quad Xiahan Shi\textsuperscript{1} \quad Volker Fischer\textsuperscript{1,}\footnotemark[1]\\
\textsuperscript{1}Bosch Center for AI (BCAI)\quad
\textsuperscript{2}Heidelberg University\quad
\textsuperscript{3}University of Freiburg
}

%\author{Elias Eulig\thanks{Corresponding authors. Contact via {\tt\small elias@eeulig.com} or {\tt\small volker.fischer@de.bosch.com}.}\quad Piyapat Saranrittichai\quad Chaithanya Kumar Mummadi\\
%Kilian Rambach\quad William Beluch \quad Xiahan Shi \quad Volker Fischer\footnotemark[1]\\
%Bosch Center for Artificial Intelligence (BCAI)
%}

%Uncomment below and comment above to see how it looks like with emails

%\author{Elias Eulig\quad Piyapat Saranrittichai\quad Chaithanya Kumar Mummadi\\
%Kilian Rambach\quad William Beluch \quad Xiahan Shi \quad Volker Fischer\\
%Bosch Center for Artificial Intelligence (BCAI)\\
%{\tt\small elias@eeulig.com,$\left\{
%\begin{tabular}{@{}c@{}}
%piyapat.saranrittichai chaithanyakumar.mummadi\\
%kilian.rambach william.beluch xiahan.shi volker.fischer
%\end{tabular}
%\right\}$@de.bosch.com}}

\maketitle

\ificcvfinal\thispagestyle{empty}\fi

\input{00_abstract}
\input{01_introduction}
\input{02_related_work}
\input{03_methods}
\input{04_baselines}

\input{05_results}
\input{06_conclusion}

\clearpage

{\small
\bibliographystyle{ieee_fullname}
\bibliography{egbib}
}

\include{appendix}

\end{document}

%% file: commands.tex
\newcolumntype{?}{!{\vrule width .2em}}

\newcommand{\commentOut}[1]{}
\definecolor{mybrown}{RGB}{165, 42, 42}
\newcommand\elias[1]{\textcolor{black}{#1}} % Todo-notes for Elias
\newcommand\volker[1]{\textcolor{black}{#1}} % Todo-notes for Volker

\newcommand\kilian[1]{\textcolor{cyan}{#1}} % Todo-notes for Kilian
\usepackage{umoline}

\usepackage{xspace}
\makeatletter
\DeclareRobustCommand\onedot{\futurelet\@let@token\@onedot}
\def\@onedot{\ifx\@let@token.\else.\null\fi\xspace}

\def\eg{e.g\onedot}

\def\wrt{w.r.t\onedot}
\makeatother

\newcommand{\figref}[1]{Fig.\ \ref{#1}}
\newcommand{\tabref}[1]{Tab.\ \ref{#1}}
\newcommand{\secref}[1]{Sec.\ \ref{#1}}
\newcommand{\figrefs}[1]{Fig.\ \labelcref{#1}}
\newcommand{\tabrefs}[1]{Tab.\ \labelcref{#1}}
\newcommand{\secrefs}[1]{Sec.\ \labelcref{#1}}

\newcommand{\diagvib}{\mbox{DiagViB-6}\xspace}
\newcommand{\ood}{OOD\xspace}
\newcommand{\position}{\texttt{position}\xspace}
\newcommand{\hue}{\texttt{hue}\xspace}
\newcommand{\lightness}{\texttt{lightness}\xspace}
\newcommand{\scale}{\texttt{scale}\xspace}
\newcommand{\shape}{\texttt{shape}\xspace}
\newcommand{\texture}{\texttt{texture}\xspace}

\newcommand\rurl[1]{%
  \href{http://#1}{\nolinkurl{#1}}%
}

%% file: 00_abstract.tex
%%%%%%%%% ABSTRACT
\begin{abstract}

%Current neural networks for object classification are known to exploit shortcut opportunities in the form of easy to extract and predictive image factors
%This is called shortcut learning as a consequence the network fails to learn more holistic decision rules and representations that enable generalization beyond the training data on out-of-distribution (o.o.d.) settings.
%While humans are also prone to shortcut learning in certain cases, the biological model appears not affected with respect to basic factors of variation (FoV) such as shape, color, or texture, and can distinguish these factors even in cases one factor has a high predictive power for another.
%We argue that both shortcut and generalization opportunities for these basic FoV are inherent properties of real-world vision data and arise from varying degrees of correlation.
%In order to overcome shortcut learning, we consider it a necessary criterion for a vision model to be able to exploit generalization opportunities present in the data.
%We argue that the presence of varying degrees of correlation among these basic FoV is an inherent property of real-world vision data and provides both shortcut and \emph{generalization opportunities}.
%We introduce a framework including diagnostic datasets and suitable metrics to evaluate a network's generalization capabilities for six independent visual object factors.

%Instead of learning holistic object representations from vision data,
Common deep neural networks (DNNs) for image classification have been shown to rely on \emph{shortcut opportunities} (SO) in the form of predictive and easy-to-represent visual factors.
This is known as shortcut learning and leads to impaired generalization.
%Shortcut learning refers to the phenomenon that deep neural networks (DNN) rely on shortcut opportunities in the form of predictive and easy-to-extract image factors, instead of learning more holistic representations that generalize to out-of-distribution (o.o.d.) settings.
In this work, we show that common DNNs also suffer from shortcut learning when predicting only basic visual object factors of variation (FoV) such as shape, color, or texture.
%This is not the case for the human model which remains largely unaffected with respect to FoVs.
%While humans are also prone to shortcut learning in some specific cases, the biological model remains largely unaffected with respect to these basic FoVs.
We argue that besides shortcut opportunities, \emph{generalization opportunities} (GO) are also an inherent part of real-world vision data and arise from partial independence between predicted classes and FoVs. 
We also argue that it is \emph{necessary} for DNNs to exploit GO to overcome shortcut learning.
Our core contribution is to introduce the Diagnostic Vision Benchmark suite \emph{DiagViB-6}, 
which includes datasets and metrics to study a network's shortcut vulnerability and generalization capability for six independent FoV. 
In particular, \emph{DiagViB-6} allows controlling the type and degree of SO and GO in a dataset.
%We consider it a \emph{necessary} prerequisite that a good vision model should exploit given generalization opportunities to improve generalization on these factors.
We benchmark a wide range of popular vision architectures and show that they can exploit GO only to a limited extent.
%We further argue that \emph{novel architectures} can provide powerful solutions to overcome this limitation and illustrate how this can be done by proposing architecture variants that, for specific factors, significantly improve over the baselines.
%To illustrate how this could be achieved, we present architecture variants that \kilianText{, for specific factors, significantly improve over the baselines.}{significantly improve over the baselines for specific factors.}

%To summarize, the contribution of this work is threefold: (1) proposing a diagnostic framework for shortcut learning; (2) evaluating the shortcut vulnerability of common vision networks; (3) proposing a superior, new architecture.
\end{abstract}

%Common deep neural networks (DNN) for image classification have been shown to rely on shortcut opportunities (SO) in the form of predictive and easy-to-represent visual factors. This is known as shortcut learning and leads to impaired generalization. In this work, we show that common DNNs also suffer from shortcut learning when predicting only basic visual object factors of variation (FoV) such as shape, color, or texture. We argue that besides shortcut opportunities, generalization opportunities (GO) are also an inherent part of real-world vision data and arise from partial independence between predicted classes and FoVs. We also argue that it is necessary for DNNs to exploit GO to overcome shortcut learning. Our core contribution is to introduce the Diagnostic Vision Benchmark suite DiagViB-6, which includes datasets and metrics to study a network's shortcut vulnerability and generalization capability for six independent FoV. In particular, DiagViB-6 allows controlling the type and degree of SO and GO in a dataset. We benchmark a wide range of popular vision architectures and show that they can exploit GO only to a limited extent.

%% file: 01_introduction.tex
\section{Introduction}
\label{sec:introduction}

Despite their state-of-the-art performance on object classification tasks, deep neural networks (DNN) are highly prone to shortcut learning \cite{geirhos2018imagenet, sauer2021counterfactual, hermann2019exploring}.
Instead of learning holistic representations and decision rules 
that can generalize beyond the training data, DNNs overly rely on so-called \emph{shortcut opportunities} (SO), which occur when the target class is highly correlated to one or very few easy-to-represent input factors \cite{hermann2020shapes}.
This leads to poor generalization on many out-of-distribution (\ood) settings, \eg ImageNet trained DNNs are biased towards texture % for the object classification task 
and fail to generalize under texture-shape cue conflict evaluation \cite{geirhos2018imagenet}.
%Similar to \cite{geirhos2020shortcut}, we consider this vulnerability one of the primary sources for poor generalization on out-of-distribution (\ood) examples, such as poor compositional generalization.
%A prominent example is the texture bias of DNNs on the ImageNet dataset \cite{geirhos2018imagenet}, where the DNNs heavily base their classification decisions on the texture, while ignoring the object shape.
\begin{figure}
    \begin{center}
    \input{figures/Figure_1}
    \vskip -\baselineskip
    \caption{\label{fig:fig1} Exemplar study in our proposed benchmark. The network is trained to predict factor classes \texttt{2}, \texttt{4}, \texttt{3} for the \shape factor with varying \hue. All five depicted training combinations are uniformly shown during training. The shape \texttt{2} co-occurs solely with the \texttt{blue} class of the \hue factor, which poses a shortcut opportunity. The shapes \texttt{4} and \texttt{3} occur uniformly with hue \texttt{red} and \texttt{green}; these combinations pose a generalization opportunity, since they reduce the predictiveness of the \hue factor for the \shape factor.
    %Note that inside the factor classes, e.g. \texttt{3} or \texttt{blue}, some variations exist, i.e., different samples for the shape \texttt{3} or different variants of \texttt{blue} can be used.
    Test accuracy is computed on examples from OOD factor combinations to evaluate a model's shortcut vulnerability in the context of the given generalization opportunity.
    }
    \end{center}
\end{figure}
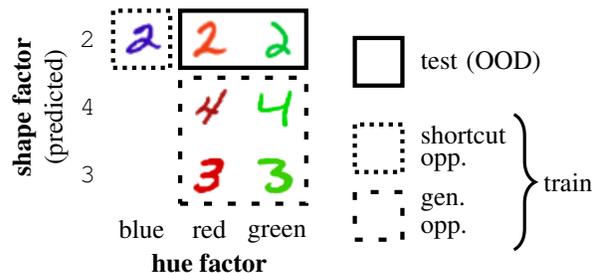
% While humans are also prone to shortcut learning in certain cases, 
% such as object classification under cue-conflict settings \cite{oliva2007role},
% the biological model remains largely unaffected by shortcuts when it comes to predicting \emph{basic} object factors of variation.
% By factors of variation (FoV), we mean basic attributes that define the visual appearance of an object, such as color, shape or texture.
% If a shortcut-vulnerable classification network overly relies on the color as a shortcut, causing it to ignore the more complicated factor shape, an image such as a blue elephant would be a troubling case.
% Humans, however, have a shortcut-robust representation of both color and shape, and can easily recognize this example as a ``blue elephant''. 
% This ``shortcut-immunity'' in handling basic FoV is just as necessary for intelligent systems. In order to act in an open-world context, the predictions of a model have to remain reasonable under changing conditions.

While humans are also prone to shortcut learning in certain cases, 
such as object classification under context-based cue conflict settings \cite{oliva2007role}, the biological model remains largely unaffected by shortcuts when it comes to predicting \emph{basic} object factors of variation (FoV) such as \shape, \hue or \texture.
This ``shortcut-immunity'' \elias{\wrt} basic FoV is just as necessary for intelligent systems; thus, efforts towards improving model generalization are \elias{of utmost importance}.
%In order to act in an open-world context, the predictions of a model have to remain consistent under changing conditions.

Existing literature studies shortcut behavior in DNNs mainly in the context of object classification. In this work, we address a more fundamental variant of shortcut learning, focusing specifically on the prediction of basic FoV themselves, similar to \cite{hermann2020shapes}.
%can exhibit shortcut learning 
%This work approaches shortcut learning in object classification from a different angle: Instead of considering the performance on the actual \emph{object classes} as done in existing literature, we study a DNN's shortcut vulnerability in learning \emph{basic factors of variation} (FoV) such as shape, color or texture.
%While excellent performance on in-distribution test sets is easily achieved, the shortcut vulnerability of these models leads to poor generalization 
In the context of FoV prediction, we refer to different manifestations of a factor as \emph{factor classes}. For example, ``red'' and ``green'' are factor classes for the factor \hue, whereas ``circle'' and ``elephant shape'' are factor classes for the factor \shape.
%Different realizations of such a FoV are labeled with certain \emph{factor classes} such as ``red'' or ``green'' for color, or ``circular'' or ``elephant shape'' for shape.
Object classes such as ``elephant'' or ``car'' are characterized by the co-occurrence of certain factor classes, e.g. ``gray'', ``elephant shape'', and ``elephant texture'' characterize an elephant. SO arise from this co-occurrence of different factor classes.
%\kumar{why shortcut learning on FoV is interesting? or What is the motivation for focusing on FoV? what additional insights do we get here when compared with object classification ?}

%Note, that while generally an image is generated by numerous FoV that could be used as a shortcut, in this work we from now on consider only FoV for objects, ignoring for example background factors.

%As stated in \cite{geirhos2020shortcut}, SO are a property of real-world vision data. In this work, we propose to additionally consider \emph{generalization opportunities} (GO). GO take the form of violations of the strict correlation between a target class and an input FoV at training time. For instance, a dotted and striped texture is not constrained to occur in particular colors. We refer to such cases as \emph{compositional-based} GO.
%which are the more natural kind of GO in real-world datasets. (No examples of proof for this statement)
As stated in \cite{geirhos2020shortcut}, SO are a property of real-world vision data. In this work, we propose to additionally consider \emph{generalization opportunities} (GO), which are a relaxation of the strict correlation (or co-occurrence) between a target class and an input FoV at training time. For instance, consider the object ``car'' with factors \shape and \texttt{color}; cars appearing in different colors during training would induce a GO compared to cars appearing only in one color.
%For instance, school buses with ``red'' and ``green'' provide GO on the dataset, which contains school buses that always appear in ``yellow''. 
We refer to such cases as \emph{compositional-based} GO. Correlations can also be violated in the form of outliers consisting of rare combinations of factor classes, e.g. a ``white elephant''. We refer to these cases as \emph{frequency-based} GO.
%A more realistic type of generalization opportunity arises in case certain factor classes are not correlated to classes of another factor. 
%For example, in real-world simple shapes such as circles and rectangles are not correlated to certain textures.
%Besides, correlations that promote GO might be underrepresented or formed with rare combinations of factor classes, e.g. a ``white elephant''. We refer to such cases as \emph{frequency-based} GO. Just like SO, GO are also an integral part of real-world data, and should be exploited to overcome shortcut learning.

%A straightforward approach resulting in more GO in the training data is data augmentation.
A straightforward approach to introduce GO in the training data is data augmentation.
For example, \cite{krizhevsky2012alexnet} applies random color transformations to the training images, removing a potential correlation between the target class and the factor color. However, data augmentation results in models that are invariant with respect to the augmented factor. Such models lose important information that may be needed
%This renders the model thus incapable 
to properly identify and reason about \ood samples.
In contrast to being invariant, we argue that a good vision model needs to have an explicit representation of these FoV.
%In this work we argue, that a general vision model has to overcome shortcut learning wrt. basic visual FoV in the presence of generalization opportunities.
Our work aims at analyzing a model's capability to exploit the GO already present in the data, as opposed to adding more GO to a dataset, as done in data augmentation.
While several synthetic benchmark datasets for compositional generalization and cue-conflict settings already exist \cite{hermann2020shapes, atzmon2020causal, Nagarajan_2018_ECCV_attributes_as_operators}, none of them enables sufficient and systematic control over the SO and GO present in the dataset for a broad set of different visual object FoV.

\begin{figure}[tb]
	\begin{center}
		\includegraphics[width=\linewidth]{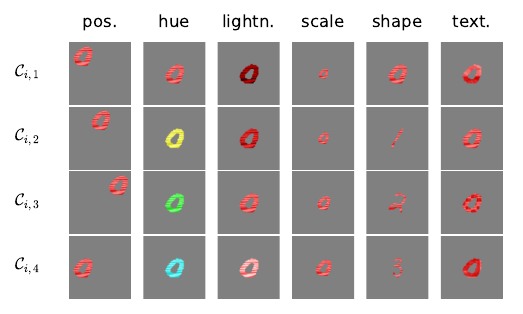}
		\vskip -0.1in
		\caption{Image space traversal across all six FoV and four corresponding class labels used in this work. Along each column only the corresponding factor is varied, while all others are fixed. Note that some factors have more than the four classes shown here (refer \tabref{tab:factors_overview}).}
		\vskip -0.1in
		\label{fig:image-space-traversal}
	\end{center}
\end{figure}

Inspired by prior work on shortcut learning and compositional generalization \cite{geirhos2020shortcut, hermann2020shapes}, we present a synthetic but diagnostic benchmark suite \diagvib\footnote{\elias{\url{https://github.com/boschresearch/diagvib-6}}}that includes different studies to evaluate a model's shortcut vulnerability %in the presence of different GO. 
under varying degrees of GO.
Figure \ref{fig:fig1} illustrates an exemplar study in our benchmark.
The benchmark suite contains an image-generating function that allows direct and independent control over the six basic, visual object FoV: \position, \hue, \lightness, \scale, \shape, and \texture (\figref{fig:image-space-traversal}). Additionally, our framework provides a dataset-generating function that enables a user to control the nature of SO and GO appearing in a dataset. This is achieved by introducing different degrees of correlation between factors, and inducing co-occurrences of certain factor class combinations.
%GO and SO are induced by correlating the appearance of certain factor classes in produced images in a controlled and structured manner.
%Furthermore, the benchmark suite also provide metrics to evaluate a model's shortcut vulnerability for each factor, as well as its generalization ability for different GO. 
Furthermore, the benchmark suite provides metrics to evaluate a model's shortcut vulnerability under different GO for each factor.

We evaluate a wide range of common deep learning vision models on our benchmark and perform an exhaustive investigation of their shortcut vulnerability \wrt the six stated FoV. We show that while they exploit frequency GO, they exploit the more relevant compositional GO only to a limited extent.
This holds true also for approaches specifically designed to counteract shortcut learning.

%Ultimately, the design of our benchmark suite allow a user to simulate and control different degree of SO and GO in a dataset (not commonly available in real world data) and also assess a model's behavior with regards to its shortcut vulnerability and generalization ability under various controlled tasks and data setups. We admit that this benchmark suite does not \emph{sufficiently} and directly endorse a vision model's generalization ability on a real world data (\eg object classification on ImageNet). However, it serves as a critical diagnosis that is \emph{necessary} to study a model's behavior under different conditions.\elias{Too vague. It is necessary to benchmark the shortcut vulnerability and generalization capability of DNNs} 

We admit that this benchmark suite does not \emph{sufficiently} and directly prove a vision model's ability to generalize on real-world data (\eg object classification on ImageNet). However, it serves as a critical diagnosis that is \emph{necessary} in order to study a model's shortcut vulnerability and generalization ability under various controlled tasks and data setups. Ultimately, the design of our benchmark suite allows a user to control different degrees of SO and GO in a dataset (not commonly available in real-world data), in order to assess a model's behavior under different conditions.

%Since our framework does not aim at analyzing the architectural mechanics or internal representations of a model, it allows to be easily applied to a large variety of different vision architectures.

Our contributions in this work can be summarized as follows: We propose a benchmark suite to create datasets that enable the user to independently combine six visual FoV, allowing explicit control over which SO and GO are present in the resulting data. We establish suitable metrics to evaluate both a model's shortcut vulnerability, and its capability to exploit GO in the data. Lastly, we provide empirical evidence that common vision architectures exploit GO only to a limited extent, especially compositional-based GO.
%We evaluate shortcut vulnerability by measuring compositional generalization capabilities in the context of different training datasets offering different versions and degrees of generalization opportunities.

%We want the network to have a complete inner representation of all object-defining FoV and argue that this is best done through architecture design.
% Image space traversal figure

%% file: figures/Figure_1.tex
\begin{tikzpicture}

\def\CW{0.9}
\def\CSize{1.8}
\def\XG{0.0}
\def\YG{0.0}
\def\LTTY{1.5}
\def\LOOY{-1.0}

% Draw frame.
%\draw (\XG + 1*\CW,\YG + 1*\CW) node [draw=black, rectangle, inner sep=0pt, minimum size=9.2*\CW em] {};

% Plot y-axis labels.
\draw (\XG - 1.5*\CW,\YG + 1*\CW) node [rotate=90] {
\begin{tabular}{c}
    \bf{shape factor} \\
    (predicted)
\end{tabular}
};
\draw (\XG - 0.8*\CW,\YG + 2*\CW) node {\texttt{2}};
\draw (\XG - 0.8*\CW,\YG + 1*\CW) node {\texttt{4}};
\draw (\XG - 0.8*\CW,\YG + 0*\CW) node {\texttt{3}};

% Plot x-axis labels.
\draw (\XG + 1*\CW,\YG - 1.3*\CW) node {\bf{hue factor}};
\draw (\XG + 0*\CW,\YG - 0.8*\CW) node {blue};
\draw (\XG + 1*\CW,\YG - 0.8*\CW) node {red};
\draw (\XG + 2*\CW,\YG - 0.87*\CW) node {green};

% Plot legend train / test.
\draw (\XG + 3.5*\CW,\YG + 0*\CW + \LTTY) node [draw=black, ultra thick, rectangle, inner sep=0pt, minimum size=2*\CW em] {};
%\draw (\XG + 3.5*\CW - 0.1,\YG + 1*\CW + \LTTY + 0.1) node [draw=black, dashed, ultra thick, rectangle, inner sep=0pt, minimum size=2*\CW em] {};
%\draw (\XG + 3.5*\CW + 0.1,\YG + 1*\CW + \LTTY - 0.1) node [draw=black, dotted, ultra thick, rectangle, inner sep=0pt, minimum size=2*\CW em] {};
%\draw (\XG + 3.5*\CW,\YG + 1*\CW + \LTTY) node {\includegraphics[width=\CSize em]{figures/F1_black.png}};

\draw (\XG + 3.5*\CW,\YG + 1*\CW + \LTTY) node [text=white] {\scalebox{2}{\(\star\)}};
\node [text width=5cm] at (\XG + 6.9*\CW,\YG + 0*\CW + \LTTY) {test (\ood)};
%\node [text width=5cm] at (\XG + 7*\CW,\YG + 1*\CW + \LTTY) {training};

% Draw SO and GO.
\draw (\XG + 0*\CW,\YG + 2*\CW) node [ultra thick, dotted, draw=black, rectangle, inner sep=0pt, minimum size=2.4*\CW em] {};
\draw (\XG + 1.5*\CW,\YG + 0.5*\CW) node [ultra thick, loosely dashed, draw=black, rectangle, inner sep=0pt, minimum size=5.3*\CW em] {};

% Draw test cases.
%\draw (\XG + 0*\CW,\YG + 0.5*\CW) node [ultra thick, draw=black, rectangle, inner sep=0pt, minimum width=2.4*\CW em, minimum height=5.3*\CW em] {};
\draw (\XG + 1.5*\CW,\YG + 2*\CW) node [ultra thick, draw=black, rectangle, inner sep=0pt, minimum width=5.3*\CW em, minimum height=2.4*\CW em] {};

% Plot legend for SO and GO.
\draw (\XG + 3.5*\CW,\YG + 1.5*\CW + \LOOY) node [ultra thick, dotted, draw=black, rectangle, inner sep=0pt, minimum size=2*\CW em] {};
\draw (\XG + 3.5*\CW,\YG + 0.5*\CW + \LOOY) node [ultra thick, loosely dashed, draw=black, rectangle, inner sep=0pt, minimum size=2*\CW em] {};
\node [text width=5cm] at (\XG + 6.9*\CW, \YG + 1.7*\CW + \LOOY) {shortcut};
\node [text width=5cm] at (\XG + 6.9*\CW, \YG + 1.3*\CW + \LOOY) {opp.};
\node [text width=5cm] at (\XG + 6.9*\CW, \YG + 0.7*\CW + \LOOY) {gen.};
\node [text width=5cm] at (\XG + 6.9*\CW, \YG + 0.3*\CW + \LOOY) {opp.};
\draw [very thick, decoration={brace,amplitude=8pt},decorate] (\XG + 5.5*\CW, \YG + 2.0*\CW + \LOOY) -- (\XG + 5.5*\CW, \YG + 0.0*\CW + \LOOY);
\node at (\XG + 6.3*\CW, \YG + 1.0*\CW + \LOOY) {train};

% 2
\draw (\XG + 0*\CW,\YG + 2*\CW) -- (\XG + 0*\CW,\YG + 2*\CW) node {\includegraphics[width=\CSize em]{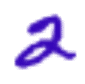}};
\draw (\XG + 1*\CW,\YG + 2*\CW) -- (\XG + 1*\CW,\YG + 2*\CW) node {\includegraphics[width=\CSize em]{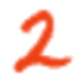}};
\draw (\XG + 2*\CW,\YG + 2*\CW) -- (\XG + 2*\CW,\YG + 2*\CW) node {\includegraphics[width=\CSize em]{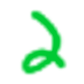}};
% 4
%\draw (\XG + 0*\CW,\YG + 1*\CW) -- (\XG + 0*\CW,\YG + 1*\CW) node {\includegraphics[width=\CSize em]{figures/F1_blue_4.png}};
\draw (\XG + 1*\CW,\YG + 1*\CW) -- (\XG + 1*\CW,\YG + 1*\CW) node {\includegraphics[width=\CSize em]{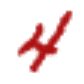}};
\draw (\XG + 2*\CW,\YG + 1*\CW) -- (\XG + 2*\CW,\YG + 1*\CW) node {\includegraphics[width=\CSize em]{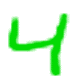}};
% 3
%\draw (\XG + 0*\CW,\YG + 0*\CW) -- (\XG + 0*\CW,\YG + 0*\CW) node {\includegraphics[width=\CSize em]{figures/F1_blue_3.png}};
\draw (\XG + 1*\CW,\YG + 0*\CW) -- (\XG + 1*\CW,\YG + 0*\CW) node {\includegraphics[width=\CSize em]{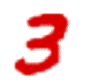}};
\draw (\XG + 2*\CW,\YG + 0*\CW) -- (\XG + 2*\CW,\YG + 0*\CW) node {\includegraphics[width=\CSize em]{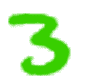}};

\end{tikzpicture}

%% file: 02_related_work.tex
\section{Related work}
\label{sec:related_work}

\paragraph{Compositional generalization in visual attribute prediction}
Our work is formulated under the general framework of visual attribute prediction. In contrast to classical object recognition, the task is to learn semantic attributes of a given object \cite{farhadi2009describing,russakovsky2010attribute}. By learning such class-agnostic visual object attributes, models can make useful predictions about object classes that have not been seen at training time.
We are interested in specific scenarios, called \textit{compositional generalization}, in which only a limited number of attribute combinations are provided during training. The model needs to generalize during training in order to handle inputs with unseen attribute combinations during testing \cite{misra-red-wine2017, Nagarajan_2018_ECCV_attributes_as_operators, Yang_2020_CVPR, Li_2020_CVPR-symmetry-group-object-attribute-pairs, Alfassy_2019_CVPR, atzmon2020causal, Yu2014_utzappos50k}. While most of these works use a multi-task setup requiring \elias{object and attribute annotations for every image} during training, we consider a single-task prediction, in which only one factor is predicted. In addition, we consider predicting more fundamental factors (\eg lightness, scale, hue) that are independent of each other and can be applied to all types of objects, as opposed to high-level visual attributes (\eg glossy, furry, smooth).

\vspace{-8pt}
\paragraph{Shortcut learning}
In recent years, 
%Recently,
multiple works have shown that DNNs are vulnerable to shortcut learning on several real-world and synthetic datasets \cite{baker2018deep,geirhos2018imagenet, hermann2019exploring,geirhos2020shortcut,hermann2020shapes}. 
%As a result, many studies investigate why such shortcut learning occurs, and propose methods to alleviate the vulnerability of DNNs to it.  
As a result, many studies investigate the reasons for shortcut learning, and propose methods to alleviate such vulnerabilities.  
%In \cite{hermann2019exploring} the authors suggested that differences between humans and DNNs may arise from differences in the data that they see. 
It is suggested in \cite{hermann2019exploring} that differences between humans and DNNs may arise from differences in the data that they see. 
\cite{hermann2020shapes} investigated how a model's representations are shaped by inductive biases in the presence of SO.
% Solution
A common attempt to overcome shortcut learning is to augment the training data with handcrafted transformations in order to reduce the importance of each individual factor (e.g. shape or texture) \cite{geirhos2018imagenet, li2020shape}.
%\cite{geirhos2018imagenet} increases shape bias by training deep networks with Stylized-ImageNet (SIN). \cite{li2020shape} proposed to augment training data by combining texture and shape across classes.
This approach has further been extended to generative model-based augmentations \cite{peebles2020, voynov2020unsupervised, sauer2021counterfactual}. 
%\cite{sauer2021counterfactual} proposed a network that explicitly models fundamental mechanisms which can improve \ood robustness.
Recently, \cite{minderer2020automatic} showed that for self-supervised learning, certain shortcut features can be removed automatically, %using self-supervised representation learning, 
under the assumption that such features are most vulnerable to adversarial attacks. 
%Motivation and difference
Most of the aforementioned works on shortcut learning are benchmarked on black-box datasets (e.g. Stylized ImageNet \cite{geirhos2018imagenet}), leading to a limited, implicit knowledge of the SO and GO introduced in both the training and test data. In contrast, our \diagvib enables explicit image generation, allowing shortcuts learned from individual FoV to be evaluated.

\vspace{-10pt}
\paragraph{Benchmarks}
Numerous works \volker{exist on} related areas, such as compositional generalization or disentanglement of FoV, that introduce datasets \volker{which allow} to control image factors to some extent \elias{\cite{hermann2020shapes,johnson2017clevr,atzmon2020causal,hyunjik2018factorvae, klinger2020study, baker2018deep,higgins17_zero_shot_transfer,leCun2004_generic_object_recognition, Isola2015_StatesAndTransformations,Yu2014_utzappos50k}}. However, these all contain shortcomings that we try to address in our work.
%These contain shortcomings that we address in our work:
%\volker{(1) more intra-factor class variation, more FoV, more factor classes, (2) architecture agnostic, (3) structured \& exhaustive evaluation of shortcut behavior w.r.t. individual factors.}

In contrast to \cite{hermann2020shapes, atzmon2020causal, hyunjik2018factorvae, klinger2020study,baker2018deep, higgins17_zero_shot_transfer}, our dataset contains a richer intra-factor class variation. 
E.g. for the \shape factor, each digit class of MNIST specifies a factor class. During image generation we use different instances of each individual factor class.
Similarly, for \textit{red}, we use different tones of red. Other datasets only provide for example a single cylinder shape or red tone. 
For an overview of the variation provided for each of our six factors see \secref{supp:factor_class_definitions}.

Unlike the datasets in \cite{hermann2020shapes, klinger2020study, baker2018deep, leCun2004_generic_object_recognition}, which use only 2-3 FoV, \diagvib includes six fundamental FoV that all good vision models should be shortcut-robust towards. \

%Moreover, \diagvib includes 6 FoV, where other datasets allow only for 2-3 FoV \cite{hermann2020shapes, klinger2020study, baker2018deep, leCun2004_generic_object_recognition}.
Some works, \eg \cite{hermann2020shapes}, investigate the internal feature representations of DNNs at certain layers, whereas our benchmark depends only on a model's predictions and is therefore architecture agnostic.

In contrast to multi-attribute prediction \cite{atzmon2020causal,Isola2015_StatesAndTransformations, Yu2014_utzappos50k}, we evaluate a model's shortcut behavior w.r.t. a single factor under different correlations to other FoV. This allows for a more \elias{structured and comprehensive} analysis of shortcut behavior, where the interaction between individual factors can be investigated.

%% file: 03_methods.tex
\section{Diagnostic vision benchmark suite}
\label{sec:methods}
%Images in real-world datasets are often assumed to be a high-dimensional representation of a lower-dimensional manifold of independent factors of variation \cite{bengio2013representation,locatello2020disentangling} \elias{(cite more)}. However, recent work has shown that correlations between those factors can facilitate shortcut learning, where models preferentially learn the factor that is easier to represent, often leading to a lack of generalization \cite{geirhos2020shortcut, hermann2020shapes}.

%Real-world datasets are not well suited to evaluate shortcut learning for the following two reasons: The datasets often contain subtle correlations between different factors, which are both difficult to detect and to control, and shortcut vulnerability can only be measured indirectly via an abstract classification task whose classes are defined by a complex combination of the initial factor classes themselves. \elias{(This description is pretty vague. Maybe cite or give some evidence / examples? It basically means. If a dog is defined as a specific shape with specific texture and only some colors (dogs are never green in real-world data) this means the task itself ``asks" for correlation in the data)}. To overcome those issues our dataset is constructed in two specific ways:
%\begin{enumerate}
%    \item Factors of variation are independent and all correlations between factors can be controlled by us.
%    \item The task is to predict the classes of those image-generating factors directly.
%\end{enumerate}

This section describes our benchmark suite \diagvib, used to examine the shortcut vulnerability and generalization capability of DNNs over six different independent FoV.  The suite consists of different image datasets tailored for different diagnostic studies, as well as suitable metrics for measuring shortcut vulnerability. We begin with an overview of the image generation process in our benchmark studies (\secref{subsec:prerequisite}), describe the benchmark setup in \secref{subsec:bench_setup}, introduce different studies in \secref{subsec:bench_studies}, and lastly establish the metrics to evaluate DNNs on our studies in Sec. \ref{subsec:bench_metrics}.

%%% SUBSECTION NOTATIONS
\subsection{Prerequisite}
\label{subsec:prerequisite}
% Factor overview table
\begin{table}[tb]
    \setlength{\tabcolsep}{3.5pt}
    \centering
    \begin{tabular}{ccccc}\toprule
    \(\mathcal{F}_i\) & \(\mathcal{S}_i\) &  \(\mathcal{N}_i\) & \(\mathcal{C}_{i,1}\) & \(\mathcal{S}_{i,1}\)\\\midrule
    position & \([0, 1]^2\) & 9 & top-left & \([0.1, 0.3]\times[0.1, 0.3]\)\\
    hue & \([0, 2\pi)\) & 6 & red & \([\ang{345}, \ang{15}]\)\\
    lightness &\([0, 1]^2\) & 4 & dark & \([0, 0.1]\times[0.4, 0.5]\)\\
    scale & \([0.69, 1.45]\) & 5 & small & \([0.69, 0.74]\)\\ 
    shape & MNIST & 10 & `0' & digits `0'\\
    texture & textures & 5 & tiles & tiles texture crops\\\bottomrule
    \end{tabular}
    \vspace{2pt}
    \caption{Overview of factors \(\mathcal{F}_i\), respective factor spaces \(\mathcal{S}_i\), and number of classes \(\mathcal{N}_i\). \(\mathcal{C}_{i,1}\) \& \(\mathcal{S}_{i,1}\) are \elias{exemplary} factor classes and factor space regions of class label 1.}
    \label{tab:factors_overview}
\end{table}
\paragraph{Fundamentals} Datasets in our benchmark suite consist of images with a single object described by a fixed, pre-defined set of six independent factors. 
Each factor \(\mathcal{F}_i,\, i \in \{1,\ldots, 6\}\) corresponds to a semantically meaningful image attribute: \shape, \texture, \hue, \lightness, \position and \scale. 
%Here \(i\) refers either a factor id \eg \(\mathcal{F}_1\) or a factor name \eg \(\mathcal{F}_{shape}\).
Every factor \(\mathcal{F}_i\) is associated with a certain \emph{factor space} \(\mathcal{S}_i\) from which factor realizations \(f_i\elias{\widehat{=} f_{\mathcal{F}_i}}\in\mathcal{S}_i\) that describe the objects are sampled. For example, lightness of an object \(f_\text{lightness}\) can be realized as a scalar sampled from \(\mathcal{S}_{\text{lightness}} = [0, 1]\). We assign \(\mathcal{N}_i\) discrete factor class labels for each factor, denoted as \(\mathcal{C}_{i,j}\), \(j \in \{1,\ldots, \mathcal{N}_i\}\),  which correspond to regions \(\mathcal{S}_{i,j} \subset \mathcal{S}_i~\forall j\) and \(\mathcal{S}_{i,j}\cap\mathcal{S}_{i,k} = \emptyset, j\neq k\). Similar to factors, each factor class \(\mathcal{C}_{i,j}\) corresponds to a semantically meaningful attribute class (\eg \(\mathcal{C}_{\text{lightness}, 1}\) refers to ``dark" and \(\mathcal{C}_{\text{hue}, 3}\) refers to ``green").
Note that our choice of classes for each factor is arbitrary and based on human intuition (similar to \cite{hermann2020shapes}) and thus our work does not evaluate a factor's general ability to act as a shortcut over another.

Table \ref{tab:factors_overview} provides an overview of the list of factors, their corresponding factor spaces, and the number of factor classes used throughout this work.  As examples, the factor spaces \(\mathcal{S}_{i,1}\) and the factor classes \(\mathcal{C}_{i,1}\) are also provided. A comprehensive overview of factor spaces for all factor classes is provided in \secref{supp:factor_class_definitions}.
\vspace{-8pt}
%%% IMAGE GENERATION
%
\paragraph{Image generation}
An image is generated by defining a combination of factor classes \(\mathcal{C}\, =\, (\mathcal{C}_{1,j}, \mathcal{C}_{2,k}, \ldots, \mathcal{C}_{6,l})\), one from each factor \(\mathcal{F}_i\).  This is done by first sampling factor realizations \(f_i\) from the corresponding regions (\(\mathcal{S}_{1,j}, \mathcal{S}_{2,k}, ..., \mathcal{S}_{6,l}\)) 
%(in this work the sampling is performed uniformly) 
and then generating unique images from those sampled \(f_i\) using an image-generating function \(I: (f_1, f_2 ..., f_6) \rightarrow \mathcal{I}\) with \(\mathcal{I} \in [0,1]^{H \times W \times 3}\), where \(H=W=128\).
%\footnote{Code for benchmark generation is available in supplementary material}.
The six FoV are incorporated into the image as follows: 
\(f_\text{shape}\) is always an instance of a digit in the MNIST dataset, which is then thresholded to yield a binary segmentation. From the two lightness values \(f_\text{lightness} = (f_\text{lightness}^{(1)}, f_\text{lightness}^{(2)})\) two colors with equal hue \(f_\text{hue}\) are generated. The \(f_\text{texture}\)'s are crops of histogram-normalized grey-scale \emph{texture} images and their pixel values serve as the coefficients of a convex combination between the two aforementioned colors, resulting in the final color.
%The mean between those two lightness values defines the overall object \emph{lightness}.
The object is then either up or down-sampled depending on the chosen \emph{scale} realization \(f_\text{scale}\), and placed at a specific \emph{position} \(f_\text{position}\) on a greyscale background. 

Figure \ref{fig:image-space-traversal} illustrates the variation over the six different FoV. 
In particular, the first column is generated with factor class labels \(\mathcal{C} = (\mathcal{C}_{pos.,j},\, \mathcal{C}_{hue,1},\, \mathcal{C}_{lightn.,3},\, \mathcal{C}_{scale,5},\\ \mathcal{C}_{shape,1}, \mathcal{C}_{text.,2})\) with \(j\in \{1,\,2,\,3,\,4\}\). Similarly, the other columns show images varying one factor while keeping the classes of all other factors fixed.
Note that the six different FoV are independent, and thus different images can be generated for each of the \(\prod_i\mathcal{N}_i = 54000\) different factor class combinations.
%%% SUBSECTION STUDIES
%
%
\subsection{Benchmark Setup}
\label{subsec:bench_setup}
In this section we lay out the fundamentals used in \secref{subsec:bench_studies} to design and conduct studies analyzing different aspects of a DNN's shortcut behavior.

%To evaluate these behaviours factor-specific, each study is comprised of single experiments for each factor combination there is.

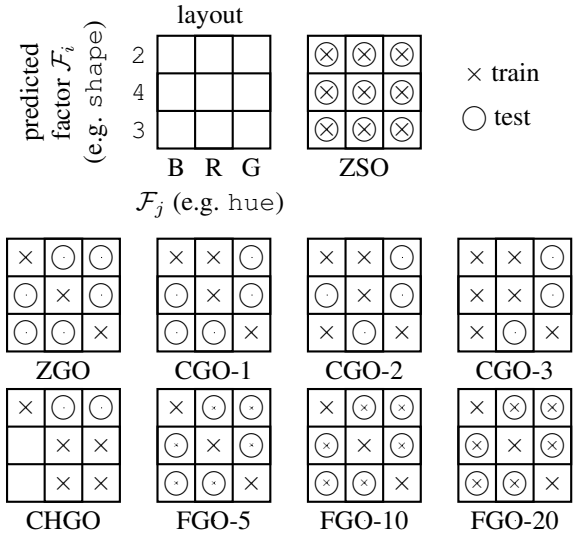
\begin{figure}
    \begin{center}
    \input{figures/studies_overview}
    \caption{\label{fig:studies_overview} Schematic overview of conducted studies (cf. \figref{fig:fig1}). For a combination of two factors \(\mathcal{F}_i, \mathcal{F}_j\) (\(x\)- and \(y\)-axis), we select three factor classes (rows and columns) for each factor and predict the first (\(y\)-axis) factor. Different selections from the resulting nine combinations of factor classes for training and test datasets (\(\times\) and \(\bigcirc\)) yield different GO and SO settings: The zero SO (ZSO), the three degrees of frequency GO (FGO) (increasing size of crosses \raisebox{0.5ex}{\scalebox{0.3}{\(\times\)}}, \raisebox{0.4ex}{\scalebox{0.5}{\(\times\)}}, \raisebox{0.3ex}{\scalebox{0.7}{\(\times\)}} indicate increasing \ood frequency during training), the zero GO (ZGO), the three compositional GO (CGO-1/2/3) settings and the compositional hybrid GO (CHGO).}
    \end{center}
    \vskip -\baselineskip
\end{figure}

For all studies, we select a subset of three factor classes from the \(\mathcal{N}_i\) available for each factor \(\mathcal{F}_i\), which are then used during training, validation and testing.
For example, as depicted in \figrefs{fig:fig1,fig:studies_overview}, we select the factor classes \(\{\text{Blue}, \text{Red}, \text{Green}\}\) and \(\{\texttt{2},\,\texttt{4},\, \texttt{3}\}\) for factors \(\mathcal{F}_{hue}\) and \(\mathcal{F}_{shape}\), respectively, and similarly for the other factors.
This results in \(3^6=729\) different factor class combinations out of the mentioned \(54000\) being used for a single experiment.
Choosing the same number of factor classes across factors simplifies evaluation and removes the number of factor classes as a source of variation.
To account for randomness introduced by this selection of factor classes, we draw five random \emph{dataset samples} of selected factor classes, over which reported results are averaged.

As mentioned in \secref{sec:introduction}, we want to evaluate a network's shortcut behavior when predicting an FoV in the presence of varying amounts of SO and GO in the training data.
In this work, we consider the interplay between \emph{all possible} pairings of two factors \((\mathcal{F}_i, \mathcal{F}_j)\).
For each pairing, the task is to predict the class of the first factor \(\mathcal{F}_i\),
where SO and GO are induced by a specified co-occurrence pattern of factor class combinations for \(\mathcal{F}_i\) and \(\mathcal{F}_j\).
For example, consider the setup in \figref{fig:fig1}, with \(\mathcal{F}_i\): \shape and \(\mathcal{F}_j\): \hue. Only five out of the nine possible combinations of factor classes (dashed rectangles in \figref{fig:fig1}) are shown during training, which induces a shortcut opportunity for the blue ``\num{2}". During testing, we then evaluate the networks performance on the \ood combinations (solid rectangles in \figref{fig:fig1}).
The other \(4\) factors are not correlated with \(\mathcal{F}_i\) or \(\mathcal{F}_j\); their factor classes all appear with the same uniform probability. We conduct an exhaustive analysis of all possible factor pairings, resulting in \(6\times(6-1)=30\) different settings, 
i.e. \(6-1\) possible pairings for each target factor.

We generate validation data following the same distribution as the training data.
In contrast, test data is designed to analyze a model's shortcut behaviour by violating the pairwise correlations existing in the training data, and thus contain \ood samples in the \(3\times 3\) matrix (cf. \(\times\) for training and \(\bigcirc\) for testing in \figref{fig:studies_overview}). 
We always generate \num{43740} training, \num{8748} validation, and \num{10000} test samples.

\subsection{Benchmark Studies}
\label{subsec:bench_studies}
%\kumar{I revised Section 3.1 and 3.2. I will now revise Section 3.3}\\
Having introduced the general structure of our benchmark, 
we now provide an overview of the five studies it enables
(see \figref{fig:studies_overview} for illustration).
%\subsubsection{Zero shortcut opportunities (ZSO)}
\vspace{-8pt}
\paragraph{Zero Shortcut Opportunities (ZSO)}
%The primary focus of our benchmark is to examine DNNs in the presence of SO and GO. Prior to the study of SO,  
The goal of this first study is to measure a network's factor classification performance in the absence of any SO. 
In the ZSO training data, each class of the target factor $\mathcal{F}_i$ 
co-occurs uniformly with all possible classes of $\mathcal{F}_j$, 
as indicated by the positions of $\times$s in the top-right of \figref{fig:studies_overview}. 
Thus, $\mathcal{F}_j$ is not predictive for $\mathcal{F}_i, ~\forall i\neq j$.
%This study only one dataset which the DNNs are trained on to predict either one of the six FoV.
%Therefore, this study does not constitute any datasets with pairwise correlations between factors.
After training, the factor-classification performance of the model is tested on a dataset that shares the same distribution as the training data (see $\bigcirc$ in \figref{fig:studies_overview}).

% \begin{itemize}
%     \item This is what we called Base Accuracy (`study\_0-1-6') before.
%     \item Motivation for this study: Our choice of classes \(\Gamma_{i,j}\) (see sec.\ \ref{subsec:notations}) is completely arbitrary and thus some classes may be harder to predict than others.
%     \item Motivation for this study: We need to know the upper bound. How well can we predict a certain factor if no shortcuts are provided and the network really has to predict that factor.
% \end{itemize}

%\subsubsection{Zero generalization opportunities (ZGO)}
\vspace{-8pt}
\paragraph{Zero Generalization Opportunities (ZGO)}
This study can be considered an ``opposite'' of the ZSO study. 
In the ZGO training data, 
the target FoV $\mathcal{F}_i$ is perfectly correlated with $\mathcal{F}_j$,
i.e. each class of $\mathcal{F}_i$ can only co-occur exclusively with one particular class of  $\mathcal{F}_j$ (see ZGO in \figref{fig:studies_overview}).
Since both factors contain redundant information for the prediction task,
$\mathcal{F}_j$ can be exploited as a shortcut to predict $\mathcal{F}_i$.
% \begin{itemize}
%     \item This is what we called full bias / full correlation (`study\_2-1-6\_100') before
%     \item Motivation: This is sort of the works case scenario of no GO and only SO and provides a per-se order for the factors and their relative shortcut vulnerability.
%     \item We correlate two factors with each other and predict one of them. So during training each class of the factor that we predicted is only shown with one specific class of the correlated factor. This way, the network has zero generalization opportunities. We then test on what happens when we violate those correlations (\ood).
% \end{itemize}

%\subsubsection{Compositional-based generalization opportunities (CGO)}
\vspace{-8pt}
\paragraph{Compositional Generalization Opportunities (CGO)}
%\volker{this could be shortened, maybe refer to intro, where comp-based GO was already addressed. the example here should be in the context of our dataset, the general example was given in intro}
%The  CGO-\(c\) studies can be seen as an interpolation between the extreme cases ZSO and ZGO.
With the ZSO and ZGO studies %in place to 
capturing the two extremes,
we continue with more realistic setups where both SO and GO are present in the training data.
As mentioned in \secref{sec:introduction}, one way to introduce GO is through compositions.
The CGO study does exactly this, addressing compositional GO by successively adding more %and more 
factor class combinations for $\mathcal{F}_i,~\mathcal{F}_j$ into the training set.
%while both the aforementioned studies provide empirical upper and lower bounds of the shortcut vulnerability of DNNs, real-world data is unlikely to contain only SO \emph{or} GO exclusively (\eg CIFAR-10, ImageNet). Instead, certain combinations of factor classes appear in the training data (\eg elephant with its texture) while other combinations remain novel for the network (\eg elephant with cat texture). 
In particular, we generate three sub-studies CGO-\(c,~c\in\{1,2,3\}\)
with increasing degree of GO  (\figref{fig:studies_overview}); the added combination is picked at random.
%where the degree of GO increases by adding new (random) pairwise factor class combinations in the training data with increasing \(c\) (see \figref{fig:studies_overview}).
%refers to the number of   (and thus GO) added during training. 
%While the pairwise correlation added in each sub-study is chosen randomly,
%the number of GO per predicted factor class is limited (per row in Fig.\ \ref{fig:studies_overview}) to two,
For each class of the target factor, we hold out at least one unseen combination for testing,
as indicated by the $\bigcirc$ symbols in each row in the CGO-\(c\) diagrams of Fig.\ \ref{fig:studies_overview}.
For \(c=0\), no GO are provided and thus CGO-0 is identical to the ZGO study discussed above. 
%we examine the behavior of DNNs on such compositional generalization tasks with varying degree of correlations provided during training.
The goal of the CGO studies is to quantify a model's generalization capability,
i.e. the capability to efficiently exploit the GO present in the data.
We would expect a good vision model to perform well on these generalization tasks,
especially as the number of SO decreases and the number of GO increases. 
% \begin{itemize}
%     \item This is what we called compositional generalization before
%     \item The study starts with the ZGO setting. In CGO-1, we then randomly add one additional combination (keep the old one as well). In CGO-2 we then randomly add another combination. The only constraint is, that there are no more than two combinations per row (Fig.\ \ref{fig:studies_overview}). This means, every class of the predicted factor $F_i$ must still have one class of the correlated factor $F_j$, that we have not seen it with during training. Otherwise there wouldn't be any \ood examples for the test case for this class.
%     \item Motivation: somewhat realistic setting (relative to frequency based) of additional factor combinations during training that is similar to GO in real-world data.
%     \item Motivation: This type of GO allows scaling / controlling the degree of GO.
%     \item Which combinations are randomly added is also randomized over the five dataset samples to average out that some combinations might be more helpful and others may be less helpul to generalize.
% \end{itemize}

%\subsubsection{Compositional-based hybrid generalization opportunities (CHGO)}
\vspace{-8pt}
\paragraph{Compositional Hybrid GO (CHGO)}
This is a special case of CGO-2, %the CGO-2 study,
for which both SO and GO are present, but explicitly separated,
as depicted in \figref{fig:studies_overview} (bottom-left).
One can think of this as a mixture of ZSO and ZGO,
in which one of the classes of $\mathcal{F}_i$ is exclusively coupled with a certain class of $\mathcal{F}_j$ (ZGO), while the remaining two target factor classes co-occur uniformly with the remaining two classes of $\mathcal{F}_j$ (ZSO). The goal of CHGO is to examine whether a model can become immune to an extreme SO by exploiting GO available for other classes of the same factors.

As pointed out in \secref{sec:introduction}, this heterogeneous mixture of SO and GO comes closest to real-world data. For instance, school buses might always be ``yellow" while other cars appear in arbitrary colors \elias{besides yellow}. Considering a model that predicts object shape, the yellow buses allow the model to shortcut its bus-shape prediction by solely relying on object color. A good vision model should now be able to exploit the provided GO, in the form of differently colored cars, to also predict the correct shape for an \ood ``red" bus.

%that examines whether factor classes learned under GO using different pairwise combinations during training generalize to unseen combinations during testing. 
%Note that the factors that compose these unseen combinations also appear under other combinations during training. 
%A depiction of this study is seen in \figref{fig:studies_overview} and also illustrated with an example in \figref{fig:fig1}.
%A special case of the CGO-2 study is when one of the predicted factor classes is presented in one single combination with one other class of the correlated factor, while the two remaining predicted factor classes are presented in both combinations with the remaining two classes of the correlated factor. We evaluate this special case separately to the CGO study in order to have a more structured insight into a models ability to exploit GO for some classes and transfer this generalization knowledge to classes for which only SO are provided at training time.
% \begin{itemize}
%     \item This is what we called the hybrid study before
%     \item Motivation: relative to the CGO, this is a more structured / explicit realization of SO and GO. Both are clearly separated here and its clear for which factor class the shortcut may happens.
% \end{itemize}

%\subsubsection{Frequency-based generalization opportunities (FGO)}
\vspace{-8pt}
\paragraph{Frequency-based Generalization Opportunities (FGO)}
Orthogonal to the CGO study, GO may also be introduced in a frequency-based manner i.e., controlling the proportion of a correlation introduced in the training data \elias{(cf.\ \cite{hermann2020shapes,beery2018terra})}. These sub-studies can also be seen as a transition from ZGO to ZSO by gradually increasing the frequency of correlations in the training data. 
%Previous studies on both synthetic \cite{hermann2020shapes} and real-world data \cite{beery2018terra} suggest that DNNs fail to exploit such GO and focus on the easier to represent yet less predictive factor instead \volker{I would remove / shorten this sentence or move to intro}.
We generate three sub-studies FGO-\(f,~f=\{5,\,10,\,20\}\), where the strict correlation from the ZGO is relaxed by means of low frequency violations (in \(f\,\si{\percent}\) of the samples) of this combination during training. Unlike CGO, all combinations are seen during training; we test on those combinations that are underrepresented.
\begin{figure*}[!t]
\centering
\begin{subfigure}{\linewidth}
	\includegraphics[width=\linewidth]{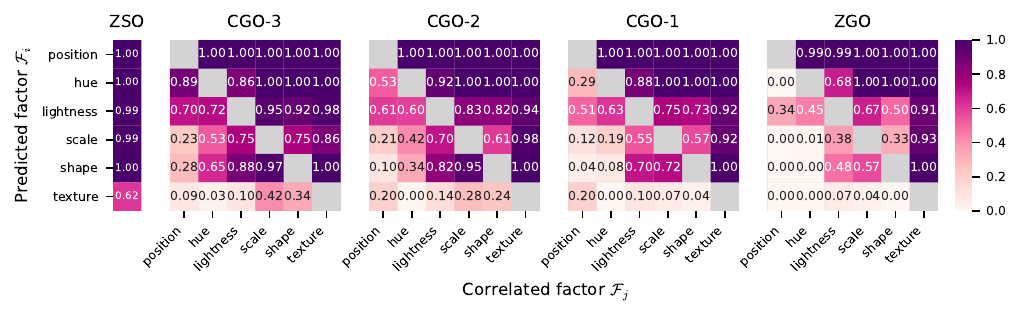}
	\vskip -0.32\linewidth
	\caption{}
	\vskip 0.27\linewidth
	\label{fig:cgo_rn18}
\end{subfigure}
\begin{subfigure}{\linewidth}
	\includegraphics[width=\linewidth]{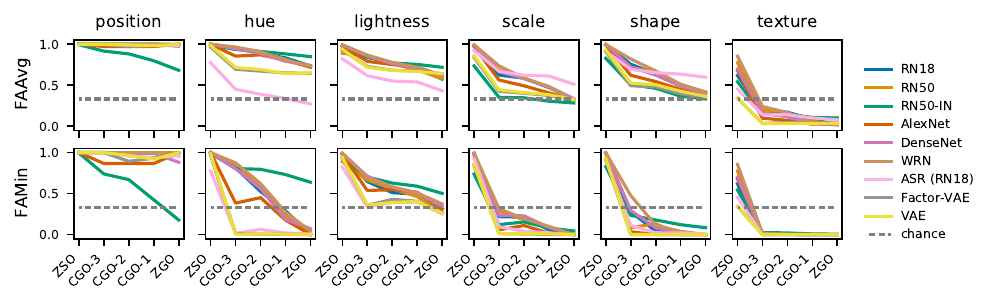}
	\vskip -0.32\linewidth
	\caption{}
	\vskip 0.27\linewidth
	\label{fig:cgo_baseline_comparison}
\end{subfigure}
\caption{Best viewed in color. (a) For the RN18 baseline, we show mean accuracies \(\text{P}_i\) for all factors \(\mathcal{F}_i\) on the ZSO study and mean accuracies \(\text{P}_{i,j}\) for all factor pairings \((\mathcal{F}_i,\mathcal{F}_j), i\neq j\) on the CGO-c and ZGO studies. (b) Aggregating model behavior over the studies, we compare all baselines using our benchmark metrics FAAvg and FAMin. Note that in (b) error bars are omitted to improve visibility of individual baselines; standard errors are reported in \secrefs{supp:zso_zgo,supp:cgo}.}
\vskip -\baselineskip
\end{figure*}
%%% SUBSECTION BENCHMARK DATASETS
%\subsection{Benchmark metrics}
\subsection{Metrics}
\label{subsec:bench_metrics}
We benchmark a model on the studies described above, starting with evaluating the mean per-class accuracy \(\text{acc}_{i,j}\) for the factor pair \((\mathcal{F}_i, \mathcal{F}_j)\), \(i,j\in\{1,\ldots,6\}\), \(i\neq j\), where \(\mathcal{F}_i\) gets predicted. 
We define the prediction accuracy \(\text{P}_{i,j}\) on the test dataset in a given study as the expectation over the five corresponding dataset samples: \(\text{P}_{i,j} = \mathbb{E}\left[\text{acc}_{i,j} \right]\).
In the case of ZSO, the prediction accuracy of \(\mathcal{F}_i\) is defined as  \(\text{P}_{i} = \mathbb{E}\left[\text{acc}_{i} \right]\), with \(\text{acc}_{i}\) being the accuracy of predicting \(\mathcal{F}_i\) on a single dataset sample.
We also define two different metrics to summarize a factor's shortcut vulnerability with respect to the other factors:

i) The Factor-Aggregated Average accuracy (FAAvg)
\vspace{-0.3em}
\begin{align}
    \text{FAAvg}_i = \mathbb{E}\left[\underset{j\in\{1,\ldots,6\}, j\neq i}{\text{mean}}\text{acc}_{i,j}\right]
\end{align}
measures the average accuracy of predicting factor \(\mathcal{F}_i\) over all possible correlations with other factors \(\mathcal{F}_j, j\in\{1,\ldots,6\}, j\neq i\). It can also be seen as a measure of the average shortcut vulnerability of the factor on a given study.

ii) The Factor-Aggregated Minimum accuracy (FAMin)
\vspace{-0.3em}
\begin{align}
    \text{FAMin}_i &= \mathbb{E}\left[\underset{j\in\{1,\ldots,6\}, j\neq i}{\text{min}}\text{acc}_{i,j}\right]
\vspace{-0.5em}
\end{align}
measures the minimum accuracy among the correlations, and can therefore be used as a measure of the maximum shortcut vulnerability of the factor on a given study. In the case of ZSO, \(\text{FAAvg}_i =\text{FAMin}_i=\text{P}_i\).

%% file: figures/studies_overview.tex
\begin{tikzpicture}

\def\CW{0.5}

\def\Xintt{4}
\def\Yintt{4.5}
\def\Xint{2.0}
\def\Yint{4.7}
\def\fcint{0.5}

\def\Xuni{4.0}
\def\Yuni{4.7}
\def\XFGO{2}
\def\YFGO{0}
\def\XFGOx{4}
\def\YFGOx{0}
\def\XFGOX{6}
\def\YFGOX{0}
\def\Xhyb{0}
\def\Yhyb{0}

\def\Xcor{0}
\def\Ycor{2}
\def\XcgI{2}
\def\YcgI{2}
\def\XcgII{4}
\def\YcgII{2}
\def\XcgIII{6}
\def\YcgIII{2}

% train / test legend.
\node [] at (6.58, 5.7) {\(\times\) train};
\node [] at (6.5, 5.1) {\(\bigcirc\) test};

\draw (\Xint - 1.5 + \CW*0.5,\Yint + \CW*1 + \CW*0.5) -- (\Xint - 1.5 + \CW*0.5,\Yint + \CW*1 + \CW*0.5) node [rotate=90] {\begin{tabular}{c}
    predicted \\
    factor \(\mathcal{F}_i\) \\
    (e.g. \shape)
\end{tabular}
};
%\draw [->, very thick] (\Xintt - 1.8 + \CW*0.5, \Yintt) -- (\Xintt - 1.8 + \CW*0.5, \Yintt + 3 * \CW);
%\draw [->, very thick] (\Xintt - 1.8 + \CW*0.5, \Yintt) -- (\Xintt - 1.8 + 3*\CW + 0.5*\CW, \Yintt);
\draw (\Xint - 0.3 + \CW*2,\Yint - 1.0 + \CW*0.5) -- (\Xint - 0.3+ \CW*2,\Yint - 1.0 + \CW*0.5) node {\(\mathcal{F}_j\) (e.g. \hue)};
\draw (\Xint - 0.3 + \CW*2,\Yint + 1.5 + \CW*0.5) -- (\Xint - 0.3+ \CW*2,\Yint + 1.5 + \CW*0.5) node {layout};

% Draw template.
\draw[thick] (\Xint, \Yint) rectangle (\Xint + \CW * 3, \Yint + \CW * 3);
\draw[thick] (\Xint + \CW, \Yint) rectangle (\Xint + \CW * 3 - \CW, \Yint + \CW * 3);
\draw[thick] (\Xint, \Yint + \CW) rectangle (\Xint + \CW * 3, \Yint + \CW * 3 - \CW);
\draw (\Xint - \fcint + \CW*0.5,\Yint + \CW*1 + \CW*0.5) -- (\Xint - \fcint + \CW*0.5,\Yint + \CW*1 + \CW*0.5) node {\texttt{4}};
\draw (\Xint - \fcint + \CW*0.5,\Yint + \CW*2 + \CW*0.5) -- (\Xint - \fcint + \CW*0.5,\Yint + \CW*2 + \CW*0.5) node {\texttt{2}};
\draw (\Xint - \fcint + \CW*0.5,\Yint + \CW*0 + \CW*0.5) -- (\Xint - \fcint + \CW*0.5,\Yint + \CW*0 + \CW*0.5) node {\texttt{3}};
\draw (\Xint + \CW*0.5,\Yint - 0.5 + \CW*0.5) -- (\Xint + \CW*0.5,\Yint - 0.5 + \CW*0.5) node {B};
\draw (\Xint + 1*\CW + \CW*0.5,\Yint - 0.5 + \CW*0.5) -- (\Xint + 1*\CW + \CW*0.5,\Yint - 0.5 + \CW*0.5) node {R};
\draw (\Xint + 2*\CW + \CW*0.5,\Yint - 0.5 + \CW*0.5) -- (\Xint + 2*\CW + \CW*0.5,\Yint - 0.5 + \CW*0.5) node {G};

% Draw the uniform setting.
\draw[thick] (\Xuni, \Yuni) rectangle (\Xuni + \CW * 3, \Yuni + \CW * 3);
\draw[thick] (\Xuni + \CW, \Yuni) rectangle (\Xuni + \CW * 3 - \CW, \Yuni + \CW * 3);
\draw[thick] (\Xuni, \Yuni + \CW) rectangle (\Xuni + \CW * 3, \Yuni + \CW * 3 - \CW);
\foreach \x in {0,1,2}
    \foreach \y in {0,1,2}
    {
        \draw (\Xuni + \CW*\x + \CW*0.5,\Yuni + \CW*\y + \CW*0.5) -- (\Xuni + \CW*\x + \CW*0.5,\Yuni + \CW*\y + \CW*0.5) node {$\times$};
        \draw (\Xuni + \CW*\x + \CW*0.5,\Yuni + \CW*\y + \CW*0.5) -- (\Xuni + \CW*\x + \CW*0.5,\Yuni + \CW*\y + \CW*0.5) node {$\bigcirc$};
    }
\draw (\Xuni + \CW + \CW*0.5,\Yuni + \CW*0 - \CW*0.5) -- (\Xuni + \CW + \CW*0.5,\Yuni + \CW*0 - \CW*0.5) node {ZSO};

% Draw the frequency setting FGO.
\draw[thick] (\XFGO, \YFGO) rectangle (\XFGO + \CW * 3, \YFGO + \CW * 3);
\draw[thick] (\XFGO + \CW, \YFGO) rectangle (\XFGO + \CW * 3 - \CW, \YFGO + \CW * 3);
\draw[thick] (\XFGO, \YFGO + \CW) rectangle (\XFGO + \CW * 3, \YFGO + \CW * 3 - \CW);
\foreach \x in {0,1,2}
    \foreach \y in {0,1,2}
{
    \ifthenelse{\x=\y}
    {
        \draw (\XFGO + \CW*\x + \CW*0.5,\YFGO + 2*\CW - \CW*\y + \CW*0.5) -- (\XFGO + \CW*\x + \CW*0.5,\YFGO + 2*\CW - \CW*\y + \CW*0.5) node {$\times$};
    }
    {
        \draw (\XFGO + \CW*\x + \CW*0.5,\YFGO + 2*\CW - \CW*\y + \CW*0.5) -- (\XFGO + \CW*\x + \CW*0.5,\YFGO + 2*\CW - \CW*\y + \CW*0.5) node {\scalebox{0.3}{\(\times\)}};
        \draw (\XFGO + \CW*\x + \CW*0.5,\YFGO + 2*\CW - \CW*\y + \CW*0.5) -- (\XFGO + \CW*\x + \CW*0.5,\YFGO + 2*\CW - \CW*\y + \CW*0.5) node {\(\bigcirc\)};
    };
}
\draw (\XFGO + \CW + \CW*0.5,\YFGO - \CW*0.5) -- (\XFGO + \CW + \CW*0.5,\YFGO - \CW*0.5) node {FGO-5};

% Draw the frequency setting FGOx.
\draw[thick] (\XFGOx, \YFGOx) rectangle (\XFGOx + \CW * 3, \YFGOx + \CW * 3);
\draw[thick] (\XFGOx + \CW, \YFGOx) rectangle (\XFGOx + \CW * 3 - \CW, \YFGOx + \CW * 3);
\draw[thick] (\XFGOx, \YFGOx + \CW) rectangle (\XFGOx + \CW * 3, \YFGOx + \CW * 3 - \CW);
\foreach \x in {0,1,2}
    \foreach \y in {0,1,2}
{
    \ifthenelse{\x=\y}
    {
        \draw (\XFGOx + \CW*\x + \CW*0.5,\YFGOx + 2*\CW - \CW*\y + \CW*0.5) -- (\XFGOx + \CW*\x + \CW*0.5,\YFGOx + 2*\CW - \CW*\y + \CW*0.5) node {$\times$};
    }
    {
        \draw (\XFGOx + \CW*\x + \CW*0.5,\YFGOx + 2*\CW - \CW*\y + \CW*0.5) -- (\XFGOx + \CW*\x + \CW*0.5,\YFGOx + 2*\CW - \CW*\y + \CW*0.5) node {\scalebox{0.5}{\(\times\)}};
        \draw (\XFGOx + \CW*\x + \CW*0.5,\YFGOx + 2*\CW - \CW*\y + \CW*0.5) -- (\XFGOx + \CW*\x + \CW*0.5,\YFGOx + 2*\CW - \CW*\y + \CW*0.5) node {\(\bigcirc\)};
    };
}
\draw (\XFGOx + \CW + \CW*0.5,\YFGOx - \CW*0.5) -- (\XFGOx + \CW + \CW*0.5,\YFGOx - \CW*0.5) node {FGO-10};

% Draw the frequency setting FGOX.
\draw[thick] (\XFGOX, \YFGOX) rectangle (\XFGOX + \CW * 3, \YFGOX + \CW * 3);
\draw[thick] (\XFGOX + \CW, \YFGOX) rectangle (\XFGOX + \CW * 3 - \CW, \YFGOX + \CW * 3);
\draw[thick] (\XFGOX, \YFGOX + \CW) rectangle (\XFGOX + \CW * 3, \YFGOX + \CW * 3 - \CW);
\foreach \x in {0,1,2}
    \foreach \y in {0,1,2}
{
    \ifthenelse{\x=\y}
    {
        \draw (\XFGOX + \CW*\x + \CW*0.5,\YFGOX + 2*\CW - \CW*\y + \CW*0.5) -- (\XFGOX + \CW*\x + \CW*0.5,\YFGOX + 2*\CW - \CW*\y + \CW*0.5) node {$\times$};
    }
    {
        \draw (\XFGOX + \CW*\x + \CW*0.5,\YFGOX + 2*\CW - \CW*\y + \CW*0.5) -- (\XFGOX + \CW*\x + \CW*0.5,\YFGOX + 2*\CW - \CW*\y + \CW*0.5) node {\scalebox{0.7}{\(\times\)}};
        \draw (\XFGOX + \CW*\x + \CW*0.5,\YFGOX + 2*\CW - \CW*\y + \CW*0.5) -- (\XFGOX + \CW*\x + \CW*0.5,\YFGOX + 2*\CW - \CW*\y + \CW*0.5) node {\(\bigcirc\)};
    };
}
\draw (\XFGOX + \CW + \CW*0.5,\YFGOX - \CW*0.5) -- (\XFGOX + \CW + \CW*0.5,\YFGOX - \CW*0.5) node {FGO-20};

% Draw the full correlation setting.
\draw[thick] (\Xcor, \Ycor) rectangle (\Xcor + \CW * 3, \Ycor + \CW * 3);
\draw[thick] (\Xcor + \CW, \Ycor) rectangle (\Xcor + \CW * 3 - \CW, \Ycor + \CW * 3);
\draw[thick] (\Xcor, \Ycor + \CW) rectangle (\Xcor + \CW * 3, \Ycor + \CW * 3 - \CW);
\foreach \x in {0,1,2}
    \foreach \y in {0,1,2}
{
    \ifthenelse{\x=\y}
    {
        \draw (\Xcor + \CW*\x + \CW*0.5,\Ycor + 2*\CW - \CW*\y + \CW*0.5) -- (\Xcor + \CW*\x + \CW*0.5,\Ycor + 2*\CW - \CW*\y + \CW*0.5) node {$\times$}
    }
    {
        \draw (\Xcor + \CW*\x + \CW*0.5,\Ycor + 2*\CW - \CW*\y + \CW*0.5) -- (\Xcor + \CW*\x + \CW*0.5,\Ycor + 2*\CW - \CW*\y + \CW*0.5) node {$\bigcirc$}
    };
}
\draw (\Xcor + \CW + \CW*0.5,\Ycor - \CW*0.5) -- (\Xcor + \CW + \CW*0.5,\Ycor - \CW*0.5) node {ZGO};

% Draw the hybrid setting.
\draw[thick] (\Xhyb, \Yhyb) rectangle (\Xhyb + \CW * 3, \Yhyb + \CW * 3);
\draw[thick] (\Xhyb + \CW, \Yhyb) rectangle (\Xhyb + \CW * 3 - \CW, \Yhyb + \CW * 3);
\draw[thick] (\Xhyb, \Yhyb + \CW) rectangle (\Xhyb + \CW * 3, \Yhyb + \CW * 3 - \CW);
\foreach \x in {1,2}
    \foreach \y in {0,1}
        \draw (\Xhyb + \CW*\x + \CW*0.5,\Yhyb + \CW*\y + \CW*0.5) -- (\Xhyb + \CW*\x + \CW*0.5,\Yhyb + \CW*\y + \CW*0.5) node {$\times$};
\draw (\Xhyb + \CW*0.5,\Yhyb + \CW*2 + \CW*0.5) -- (\Xhyb + \CW*0.5,\Yhyb + \CW*2 + \CW*0.5) node {$\times$};
\draw (\Xhyb + 1*\CW + \CW*0.5,\Yhyb + \CW*2 + \CW*0.5) -- (\Xhyb + 1*\CW + \CW*0.5,\Yhyb + \CW*2 + \CW*0.5) node {$\bigcirc$};
\draw (\Xhyb + 2*\CW + \CW*0.5,\Yhyb + \CW*2 + \CW*0.5) -- (\Xhyb + 2*\CW + \CW*0.5,\Yhyb + \CW*2 + \CW*0.5) node {$\bigcirc$};
%\draw (\Xhyb + 0*\CW + \CW*0.5,\Yhyb + 0*\CW + \CW*0.5) -- (\Xhyb + 0*\CW + \CW*0.5,\Yhyb + 0*\CW + \CW*0.5) node {$\bigcirc$};
%\draw (\Xhyb + 0*\CW + \CW*0.5,\Yhyb + 1*\CW + \CW*0.5) -- (\Xhyb + 0*\CW + \CW*0.5,\Yhyb + 1*\CW + \CW*0.5) node {$\bigcirc$};
\draw (\Xhyb + \CW + \CW*0.5,\Yhyb - \CW*0.5) -- (\Xhyb + \CW + \CW*0.5,\Yhyb - \CW*0.5) node {CHGO};

% Draw the compositional generalization 1 setting.
\draw[thick] (\XcgI, \YcgI) rectangle (\XcgI + \CW * 3, \YcgI + \CW * 3);
\draw[thick] (\XcgI + \CW, \YcgI) rectangle (\XcgI + \CW * 3 - \CW, \YcgI + \CW * 3);
\draw[thick] (\XcgI, \YcgI + \CW) rectangle (\XcgI + \CW * 3, \YcgI + \CW * 3 - \CW);
\foreach \y in {0,1,2}
    \draw (\XcgI + \CW * \y + \CW*0.5,\YcgI + 2*\CW - \CW*\y + \CW*0.5) -- (\XcgI + \CW * \y + \CW*0.5,\YcgI + 2*\CW - \CW*\y + \CW*0.5) node {$\times$};
\draw (\XcgI + \CW*1 + \CW*0.5,\YcgI + \CW*2 + \CW*0.5) -- (\XcgI + \CW*1 + \CW*0.5,\YcgI + \CW*2 + \CW*0.5) node {$\times$};
% Draw test settings.
\draw (\XcgI + \CW*2 + \CW*0.5,\YcgI + \CW*2 + \CW*0.5) -- (\XcgI + \CW*2 + \CW*0.5,\YcgI + \CW*2 + \CW*0.5) node {$\bigcirc$};
\draw (\XcgI + \CW*0 + \CW*0.5,\YcgI + \CW*1 + \CW*0.5) -- (\XcgI + \CW*0 + \CW*0.5,\YcgI + \CW*1 + \CW*0.5) node {$\bigcirc$};
\draw (\XcgI + \CW*2 + \CW*0.5,\YcgI + \CW*1 + \CW*0.5) -- (\XcgI + \CW*2 + \CW*0.5,\YcgI + \CW*1 + \CW*0.5) node {$\bigcirc$};
\draw (\XcgI + \CW*0 + \CW*0.5,\YcgI + \CW*0 + \CW*0.5) -- (\XcgI + \CW*0 + \CW*0.5,\YcgI + \CW*0 + \CW*0.5) node {$\bigcirc$};
\draw (\XcgI + \CW*1 + \CW*0.5,\YcgI + \CW*0 + \CW*0.5) -- (\XcgI + \CW*1 + \CW*0.5,\YcgI + \CW*0 + \CW*0.5) node {$\bigcirc$};
% Draw label.
\draw (\XcgI + \CW + \CW*0.5,\YcgI - \CW*0.5) -- (\XcgI + \CW + \CW*0.5,\YcgI - \CW*0.5) node {CGO-1};

% Draw the compositional generalization 2 setting.
\draw[thick] (\XcgII, \YcgII) rectangle (\XcgII + \CW * 3, \YcgII + \CW * 3);
\draw[thick] (\XcgII + \CW, \YcgII) rectangle (\XcgII + \CW * 3 - \CW, \YcgII + \CW * 3);
\draw[thick] (\XcgII, \YcgII + \CW) rectangle (\XcgII + \CW * 3, \YcgII + \CW * 3 - \CW);
\foreach \y in {0,1,2}
    \draw (\XcgII + \CW * \y + \CW*0.5,\YcgII + 2*\CW - \CW*\y + \CW*0.5) -- (\XcgII + \CW * \y + \CW*0.5,\YcgII + 2*\CW - \CW*\y + \CW*0.5) node {$\times$};
\draw (\XcgII + \CW*1 + \CW*0.5,\YcgII + \CW*2 + \CW*0.5) -- (\XcgII + \CW*1 + \CW*0.5,\YcgII + \CW*2 + \CW*0.5) node {$\times$};
\draw (\XcgII + \CW*0 + \CW*0.5,\YcgII + \CW*0 + \CW*0.5) -- (\XcgII + \CW*0 + \CW*0.5,\YcgII + \CW*0 + \CW*0.5) node {$\times$};
% Draw test settings.
\draw (\XcgII + \CW*2 + \CW*0.5,\YcgII + \CW*2 + \CW*0.5) -- (\XcgII + \CW*2 + \CW*0.5,\YcgII + \CW*2 + \CW*0.5) node {$\bigcirc$};
\draw (\XcgII + \CW*0 + \CW*0.5,\YcgII + \CW*1 + \CW*0.5) -- (\XcgII + \CW*0 + \CW*0.5,\YcgII + \CW*1 + \CW*0.5) node {$\bigcirc$};
\draw (\XcgII + \CW*2 + \CW*0.5,\YcgII + \CW*1 + \CW*0.5) -- (\XcgII + \CW*2 + \CW*0.5,\YcgII + \CW*1 + \CW*0.5) node {$\bigcirc$};
\draw (\XcgII + \CW*1 + \CW*0.5,\YcgII + \CW*0 + \CW*0.5) -- (\XcgII + \CW*1 + \CW*0.5,\YcgII + \CW*0 + \CW*0.5) node {$\bigcirc$};
\draw (\XcgII + \CW + \CW*0.5,\YcgII - \CW*0.5) -- (\XcgII + \CW + \CW*0.5,\YcgII - \CW*0.5) node {CGO-2};

% Draw the compositional generalization 3 setting.
\draw[thick] (\XcgIII, \YcgIII) rectangle (\XcgIII + \CW * 3, \YcgIII + \CW * 3);
\draw[thick] (\XcgIII + \CW, \YcgIII) rectangle (\XcgIII + \CW * 3 - \CW, \YcgIII + \CW * 3);
\draw[thick] (\XcgIII, \YcgIII + \CW) rectangle (\XcgIII + \CW * 3, \YcgIII + \CW * 3 - \CW);
\foreach \y in {0,1,2}
    \draw (\XcgIII + \CW * \y + \CW*0.5,\YcgIII + 2*\CW - \CW*\y + \CW*0.5) -- (\XcgIII + \CW * \y + \CW*0.5,\YcgIII + 2*\CW - \CW*\y + \CW*0.5) node {$\times$};
\draw (\XcgIII + \CW*1 + \CW*0.5,\YcgIII + \CW*2 + \CW*0.5) -- (\XcgIII + \CW*1 + \CW*0.5,\YcgIII + \CW*2 + \CW*0.5) node {$\times$};
\draw (\XcgIII + \CW*0 + \CW*0.5,\YcgIII + \CW*0 + \CW*0.5) -- (\XcgIII + \CW*0 + \CW*0.5,\YcgIII + \CW*0 + \CW*0.5) node {$\times$};
\draw (\XcgIII + \CW*0 + \CW*0.5,\YcgIII + \CW*1 + \CW*0.5) -- (\XcgIII + \CW*0 + \CW*0.5,\YcgIII + \CW*1 + \CW*0.5) node {$\times$};
% Draw test settings.
\draw (\XcgIII + \CW*2 + \CW*0.5,\YcgIII + \CW*2 + \CW*0.5) -- (\XcgIII + \CW*2 + \CW*0.5,\YcgIII + \CW*2 + \CW*0.5) node {$\bigcirc$};
\draw (\XcgIII + \CW*2 + \CW*0.5,\YcgIII + \CW*1 + \CW*0.5) -- (\XcgIII + \CW*2 + \CW*0.5,\YcgIII + \CW*1 + \CW*0.5) node {$\bigcirc$};
\draw (\XcgIII + \CW*1 + \CW*0.5,\YcgIII + \CW*0 + \CW*0.5) -- (\XcgIII + \CW*1 + \CW*0.5,\YcgIII + \CW*0 + \CW*0.5) node {$\bigcirc$};
% Plot label.
\draw (\XcgIII + \CW + \CW*0.5,\YcgIII - \CW*0.5) -- (\XcgIII + \CW + \CW*0.5,\YcgIII - \CW*0.5) node {CGO-3};

\end{tikzpicture}

%% file: 04_baselines.tex
\section{Baseline setup}
\label{subsec:baselines}
This section provides an overview of the baseline network architectures we evaluate on our benchmark. We consider popular, vision-based DNN architectures (not pretrained) from PyTorch's \emph{torchvision} package \cite{paszke2019pytorch}: \textbf{AlexNet} \cite{krizhevsky2012alexnet}, \textbf{ResNet18} (RN18), \textbf{ResNet50} (RN50) \cite{he2016resnet}, \textbf{Wide ResNet50-2} (WideRN) \cite{Zagoruyko2016wideresnet}, and \textbf{DenseNet-161} \cite{huang2017densenet}. 
We also include RN50 with pretrained weights on ImageNet \cite{deng2009imagenet} (\textbf{RN50-IN}), with frozen convolutional layers and a randomly initialized fully connected layer to adapt to our benchmark. We also use the prior work \textbf{Automatic Shortcut Removal} (ASR) \cite{minderer2020automatic} as a baseline for our studies. %\paragraph{VAE and Factor-VAE}
In addition, since generative models, in particular those with interpretable factorised latent representations, are promising approaches to overcome shortcut learning on classification tasks \cite{geirhos2020shortcut}, we evaluate a standard \textbf{VAE} \cite{kingma2014vae} and a \textbf{Factor-VAE} \cite{hyunjik2018factorvae} on our benchmark. 
More details on the training and setup of baselines are provided in \secref{supp:baseline_details}.

%% file: 05_results.tex
\section{Results}
\label{sec:results}
We evaluate the baselines described in \secref{subsec:baselines} on our benchmark suite described in \secref{subsec:bench_studies}. We start with the extreme cases of zero SO (ZSO) and zero GO (ZGO), followed by the other studies that introduce compositional and frequency-based GO to the training dataset.

\vspace{-1em}
\paragraph{ZSO and ZGO}
Results for the ResNet18 baseline on both these studies are presented in \figref{fig:cgo_rn18}. A comparison of all baselines using the metrics described in \secref{subsec:bench_metrics} is provided in \figref{fig:cgo_baseline_comparison}. The factor-wise results for each baseline are presented in \secrefs{supp:zso_zgo,supp:cgo}.

The ZSO study evaluates how well a model can predict each factor in the absence of SO. This yields a single mean accuracy \(\text{P}_i\) for each factor \(\mathcal{F}_i\), visualized as a vector on the very left in \figref{fig:cgo_rn18}.
Here, RN18 achieves high accuracies  (\(\geq \SI{99}{\percent}\)) for predicting all factors except \texture, for which the mean accuracy is lower \(\SI{62(6)}{\percent}\). 
A similar behavior can be observed for most other baselines (leftmost point in each plot of \figref{fig:cgo_baseline_comparison}) with a few exceptions: VAE, Factor-VAE and ASR are worse at predicting \texture, with \(\SI{34}{\percent}\), \(\SI{34}{\percent}\), and \(\SI{45}{\percent}\) accuracy, respectively. This is likely due to a limited capability in representing high-frequency structures (\secref{supp:zso_zgo}).
In summary, considering the aforementioned exceptions, all baselines can predict all factors to a reasonable extent.

The ZGO study evaluates a model's ability to predict a factor \(\mathcal{F}_i\) when each of its three factor classes co-occurs exclusively with a certain factor class of another factor \(\mathcal{F}_j\) during training.
Each such pairing \((\mathcal{F}_i, \mathcal{F}_j)\) corresponds to a single cell in the rightmost matrix plot in \figref{fig:cgo_rn18}, where \(\mathcal{F}_i\) are the rows and \(\mathcal{F}_j\) are the columns.
The color coded value in each cell indicates mean accuracy \(\text{P}_{i,j}, ~i\neq j\) on \ood test samples, for which this correlation is violated.

For RN18, we observe that for some factor combinations (\eg \((\mathcal{F}_i, \mathcal{F}_j)\): (\shape, \texture)) the model does not exploit the provided SO, and instead preferentially learns the actual task, leading to similar mean accuracies on \ood test samples compared to the corresponding ZSO accuracies for this factor.
On the other hand, for other factor combinations (\eg (\texture, \shape)) the model does exploit the provided SO, yielding \ood accuracies close to zero thus below chance level. \texture and \position factors play a special role: While \position is exploited as a shortcut when predicting any other factor, \texture is never exploited as shortcut, but is easy to shortcut by any other factor. Interestingly, \(\text{P}_{i,j} \approx 1-\text{P}_{j,i}\) seems to hold for all factor combinations, i.e., \emph{if \(\mathcal{F}_i\) is exploited as a shortcut for predicting \(\mathcal{F}_j\), then \(\mathcal{F}_j\) is not exploited as shortcut for predicting \(\mathcal{F}_i\)}.

\volker{We argue that during training either (a) the representation gets dominated by a single factor ignoring the other or (b) is a superposition of both factors, i.e., both factors share capacity of the representation.
The extent a factor is represented depends on its overall ”difficulty”, i.e., \texture appears relatively difficult to learn, and how strongly \(\mathcal{F}_i\) and \(\mathcal{F}_j\) compete over the same capacities.
During testing, \(P_{i,j}\approx 1-P_{j,i}\) holds now for (a) but not for (b). For \lightness and \position, ZGO findings indicate (b), and a superposition of both factors is learnt, utilising different capacities.
%A model's ability to represent a factor in the presence of SO is related to its inductive bias and the ZGO study could be used to diagnose the
%efficiency of inductive biases of vision models in future work.
}

This yields a dataset-dependent ranking among factors, which we use to order the rows of plots in \figref{fig:cgo_rn18}, starting with the most shortcut-robust factor \position, and ending with the most shortcut-vulnerable factor \texture. \elias{Since a model's ability to represent a factor in the presence of SO is related to its inductive bias, the ZGO study could be used to diagnose the efficiency of inductive biases of vision models in future work.}

ZSO and ZGO findings transfer to most baselines, with a few exceptions: On RN50-IN, \position is less often exploited as shortcut; the model instead exploits \hue as shortcut when predicting \position (\secref{supp:cgo}, matrix plots).

\vspace{-1em}
\paragraph{CGO}
\begin{figure*}[!t]
\centering
\begin{subfigure}{\linewidth}
	\includegraphics[width=\linewidth]{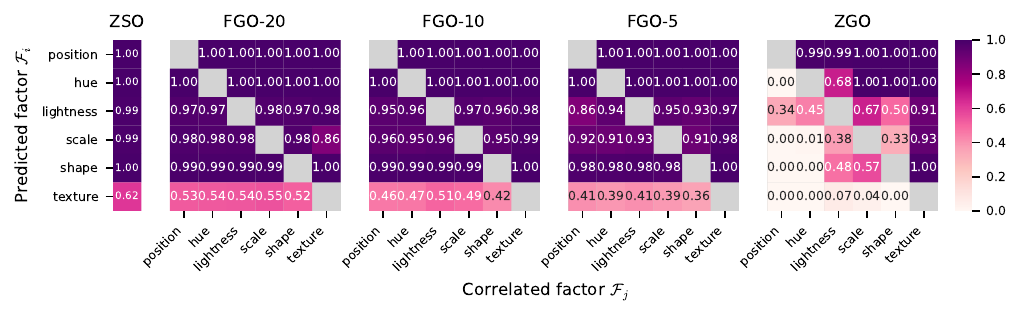}
	\vskip -0.32\linewidth
	\caption{}
	\vskip 0.27\linewidth
	\label{fig:fgo_rn18}
\end{subfigure}
\begin{subfigure}{\linewidth}
	\includegraphics[width=\linewidth]{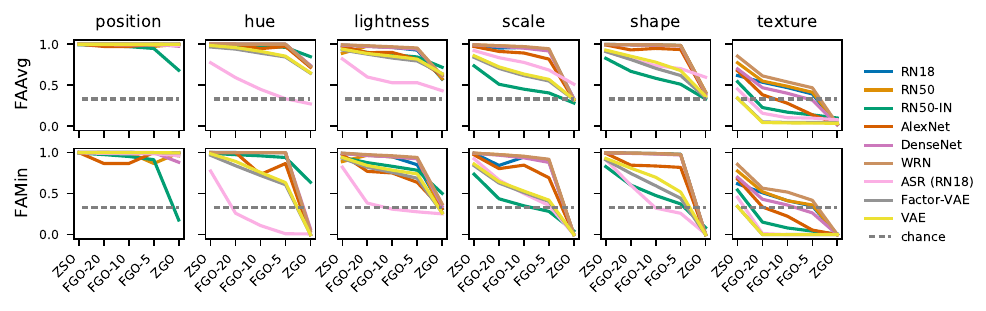}
	\vskip -0.32\linewidth
	\caption{}
	\vskip 0.27\linewidth
	\label{fig:fgo_baseline_comparison}
\end{subfigure}
\caption{Best viewed in color. Similar to \figrefs{fig:cgo_rn18,fig:cgo_baseline_comparison}, we show mean accuracies for the RN18 (a) and a comparison of all baselines (b) on the ZSO, FGO and ZGO studies. Error estimates are provided in \secrefs{supp:zso_zgo,supp:fgo}.}
\vskip -\baselineskip
\end{figure*}
We now evaluate the capability of common vision models to exploit GO provided in the form of additional factor-class combinations. 
%To this end, we generated three benchmark datasets (CGO-1,2,3) that successively add factor-class combinations to the ZGO study. 
Here, the three benchmark datasets (CGO-1,2,3) successively add factor-class combinations to the ZGO study. 
Results for RN18 are shown in \figref{fig:cgo_rn18}, and a comparison of all baselines using aggregated metrics is presented in \figref{fig:cgo_baseline_comparison}. The factor-wise results for each baseline are shown in \secref{supp:cgo}.

For RN18, we find that most factor combinations exhibit a monotonic improvement with increasing number of GO, \eg \((\scale, \hue): 0.01, 0.19, 0.42, 0.53\) for ZGO, CGO-1,2,3, respectively. However, for some factor combinations the accuracy on \ood cases does not improve with an increasing number of GO, and for some factor combinations, \eg \((\texture, \hue)\) or \((\shape, \position)\) shortcut learning occurs also for CGO-3.

This can also be seen for RN18 in the two aggregated benchmark metrics FAAvg and FAMin in \figref{fig:cgo_baseline_comparison}.
We find that FAAvg stays far below ZSO accuracy for most baselines up until three GO are added. For \texture, FAAvg is below chance level for all baselines and studies, reflecting our earlier finding that all other factors are exploited as shortcuts when predicting \texture. Moreover, FAMin stays below chance level for all baselines for \texture, \shape and \scale factors, up until three GO are added.

In contrast to \texture, for \position, none of the baselines exploit shortcuts, with RN50-IN being the only exception (\figref{fig:cgo_baseline_comparison}, leftmost column). This phenomenon is likely due to ImageNet containing limited explicit localization information \cite{he2019imagenet}, facilitating a partial invariance of the learned representations in RN50-IN.
%In contrast to \texture, the RN50-IN is the only baseline which is vulnerable to shortcut learning when predicting the \position (\figref{fig:cgo_baseline_comparison}, leftmost column). This phenomenon is likely due to ImageNet containing limited explicit localization information \cite{he2019imagenet}.\elias{I think this became misleading due to recent changes. The RN50-IN is the phenomenon, not the fact that the other baselines exploit shortcuts for the texture....}
%In contrast to \texture, only the RN50-IN is vulnerable to shortcut learning when predicting the \position - none of the other baselines exploit shortcuts for this factor.
%In contrast, only the RN50-IN is vulnerable to shortcut learning for predicting the position, and none of the other baselines exploits shortcuts for this factor. This is likely due to ImageNet containing little explicit localization information \cite{he2019imagenet}. From the CGO study, we conclude that all baselines are unable to exploit compositional-based GO for most FoV used in this work.
From the CGO study, we conclude that all baselines are unable to exploit compositional-based GO for most FoV used in this work.

\vspace{-1em}
\paragraph{CHGO}
\begin{table}
    \centering
    \fontsize{8.5}{\baselineskip}\selectfont
    \setlength{\tabcolsep}{5pt}
    \input{tables/CHGO}
	\vskip -0.1in
    \caption{Mean accuracy improvement and its standard deviation of the CHGO study over the ZGO study.}
    \label{tab:hybrid}
\end{table}
A special case of the CGO-2 study is the CHGO study, in which two classes of the predicted factor are provided with GO, while one class is presented with SO only (see \figrefs{fig:fig1,fig:studies_overview}). Comparing the accuracy on the \ood test samples for the latter with the average accuracy on the ZGO study (\tabref{tab:hybrid}), we find no improvement for most factors on all baselines, and only minor improvements for \lightness, albeit with large variance across the five dataset subsets. We conclude that all the baselines evaluated in this work do not transfer GO across classes.
\paragraph{FGO}
An orthogonal approach to providing GO through additional class-combinations during training is to relax the strict correlation between factors by means of low-frequency violations of this correlation at training time. 
We generated three benchmark datasets (FGO-\(f,~f \in \{5, 10, 20\}\)), with \(f\) indicating the frequency of the correlation violation at training time.
Results for RN18 and a comparison of all baselines are presented in \figrefs{fig:fgo_rn18,fig:fgo_baseline_comparison}. The factor-wise results for each baseline are presented in \secref{supp:fgo}.

For the RN18, the classification for all factors improves significantly with only \SI{5}{\percent} of the training samples for each class of the predicted factor violating the strict correlation from ZGO (\eg \((\shape, \hue): 0., 0.98\) for ZGO and FGO-5, respectively). When \(f\geq 10\), none of the factors except \texture are vulnerable to shortcut learning. A similar behavior can be observed for most factors and baselines (\figref{fig:cgo_baseline_comparison}), where we find that FAAvg is above chance for all factors except \texture for \(f=5\). 
%and FAMin is above chance level for all baselines and factors except \texture for \(f=10\).
However, we find that ASR fails to exploit the presented GO for \hue and \lightness when compared to other baselines, a behavior which is likely attributed to the pixel-wise reconstruction loss that is used as a regularizer in the ASR objective.
In summary, most baselines are able to exploit frequency-based GO to a certain extent on our benchmark.

%% file: tables/CHGO.tex
% ==================================================================== 
% This is an automatically generated tex file. 
% To update, please see: ./figures_and_tables/gen_table_CHGO.py 
% ==================================================================== 
\begin{tabular}{lS[table-format=2(1)]S[table-format=2(1)]S[table-format=2(1)]S[table-format=2(1)]S[table-format=2(1)]S[table-format=2(1)]S[table-format=2(1)]S[table-format=2(1)]S[table-format=2(1)]S[table-format=2(1)]S[table-format=2(1)]S[table-format=2(1)]} 
    \toprule 
     & {pos.}        & {hue}         & {lightn.}     & {scale}       & {shape}       & {text.}       \\ 
    \midrule 
            RN18 & 0(0)      & 2(3)      & 27(21)    & -1(14)    & 2(5)      & -1(2)     \\ 
            RN50 & 0(0)      & 3(5)      & 26(21)    & -1(16)    & 2(5)      & -2(2)     \\ 
         RN50-IN & -1(11)    & 5(6)      & 15(17)    & -7(11)    & 0(8)      & -1(1)     \\ 
         AlexNet & -4(8)     & 1(4)      & 24(19)    & -1(12)    & 3(6)      & -2(2)     \\
        DenseNet & 2(3)      & -3(5)     & 24(21)    & -3(13)    & 2(5)      & -1(1)     \\ 
             WRN & 1(1)      & -2(4)     & 23(22)    & 1(14)     & 3(6)      & -2(2)     \\ 
             ASR & 1(1)      & 12(15)    & 32(18)    & 1(15)     & 1(2)      & 0(2)      \\ 
           F-VAE & 0(0)      & 2(5)      & 19(17)    & -3(11)    & 1(8)      & 2(1)      \\ 
             VAE & 0(0)      & 2(5)      & 18(15)    & -3(12)    & 0(9)      & 2(1)      \\ 
    \bottomrule 
\end{tabular} 

%% file: 06_conclusion.tex
\section{Conclusion}
\label{sec:conclusion}
We addressed the lack of a suitable benchmark to evaluate the shortcut behavior of vision models on a diverse set of basic visual FoV. 
%We introduced \diagvib, a benchmark suite designed to evaluate a model's shortcut vulnerability and generalization capability, both of which are crucial performance criterion for a model. 
To this end, we introduced \diagvib, a benchmark suite designed to evaluate two crucial model performance criteria: shortcut vulnerability and generalization capability. 
Our framework allows the user to create benchmark datasets by independently combining six visual FoV, thereby precisely controlling the SO and GO present in the dataset. 
%\diagvib is architecture-agnostic, and thus applicable to a wide range of vision models.
Upon evaluation of the most commonly used vision architectures on \diagvib, we discovered that while they usually can exploit frequency-based GO, their ability to exploit compositional-based GO is limited. This finding also holds for some recent, promising approaches that have been proposed to overcome shortcut learning.

The design of \diagvib is versatile, and leads to certain natural extensions.  Three such promising future directions are: 
(a) including more complex correlation structures, potentially between more than two FoV; 
(b) transferring the benchmark design to other research areas like domain generalization and multitask learning; 
(c) extending to additional factors like background and natural corruptions.
%We strongly hope that this benchmark suite inspires and helps 
%to build truly shortcut-robust solutions for computer vision.
Lastly, we believe that our benchmark suite will inspire and help to build shortcut-robust solutions for vision models.

%% file: appendix.tex
\newpage
\clearpage
\onecolumn

\begin{center}
	\Large\textbf{DiagViB-6: A Diagnostic Benchmark Suite for 
Vision Models in the Presence of
Shortcut and Generalization Opportunities}\\
	\large\text{}\\
\author{Elias Eulig\textsuperscript{1,2}\quad Piyapat Saranrittichai\textsuperscript{1,3}\quad Chaithanya Kumar Mummadi\textsuperscript{1,3}\\
Kilian Rambach\textsuperscript{1}\quad William Beluch\textsuperscript{1} \quad Xiahan Shi\textsuperscript{1} \quad Volker Fischer\textsuperscript{1}\\
\textsuperscript{1}Bosch Center for AI (BCAI)\quad
\textsuperscript{2}Heidelberg University\quad
\textsuperscript{3}University of Freiburg
}\\

\end{center}

\renewcommand{\thetable}{A\arabic{table}}
\renewcommand{\thefigure}{A\arabic{figure}}
\setcounter{table}{0}
\setcounter{figure}{0}
\setcounter{section}{0}

\makeatletter
\afterpage{\global\setlength\@fpsep{8\p@ \@plus 2fil}}
\makeatother

\appendix
\section{Supplementary material}
\label{sec:supplementary_material}

%%% BENCHMARK DATASET SUBSECTION
\subsection{Image generation}
\label{supp:factor_class_definitions}
\begin{figure}[h!]
	\begin{center}
		\includegraphics[width=0.8\linewidth]{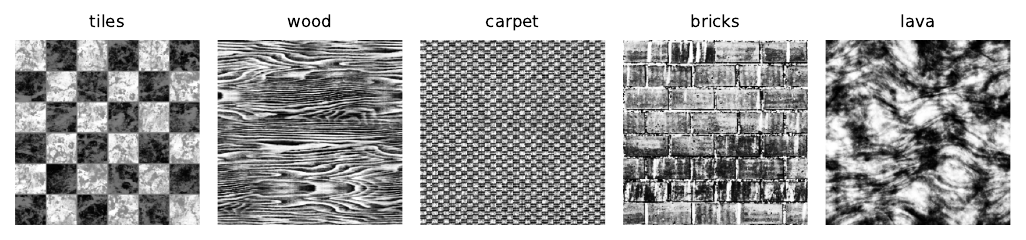}
		\caption{Overview of the textures used in this work. All textures are from \rurl{CC0Textures.com}, licensed under CC0 1.0 Universal.}
		\label{fig:textures}
	\end{center}
	\vskip -\baselineskip
\end{figure}

We provide a comprehensive overview of all factor classes and respective factor space regions in \tabref{tab:all_factors_overview}. The five textures from which the texture crops \(f_\text{text.}\) are sampled are shown in \figref{fig:textures}.

\begin{table}[h!]
    \centering
    \input{tables/all_factors_overview}
    \vspace{2pt}
    \caption{Overview of factors \(\mathcal{F}_i\), respective factor spaces \(\mathcal{S}_i\), number of classes \(\mathcal{N}_i\), factor classes  \(\mathcal{C}_{i,j}\) and respective factor space regions \(\mathcal{S}_{i,j}\) used in this work.
    }
    \label{tab:all_factors_overview}
\end{table}
%%% BASELINES SUBSECTION
\subsection{Baselines}
\label{supp:baseline_details}
We train all baseline architectures using standard cross entropy loss and Adam optimizer \cite{kingma2015adam} using early stopping. All test accuracies reported in this work correspond to the network with lowest validation loss. Batch size and learning rate were optimized on the ZSO study and weight initialization is fixed across all experiments. The output size of all the architectures are modified to predict the factor classes in our studies.

Following ASR, we use a variant of their U-Net architecture for \elias{the} lens and ResNet18 as feature extractor. Instead of an auxiliary task, we train both the lens and feature extractor using the classification task directly. \elias{The} lens is trained using least likely adversarial loss using the classification objective and also the reconstruction loss, both with equal weighting. Output of \elias{lens} is used as inputs for training the feature extractor.

Both VAE methods use architectures similar to \cite{locatello2019challenging} with a latent size of \(12\). We train both VAE with batch size \(64\) and Adam optimizer \cite{kingma2015adam} with learning rate \(0.0001\). For the Factor-VAE, we use $\gamma=20$ and train the discriminator with learning rate \(0.00001\) throughout our experiments. Some examples of latent traversals from VAE and Factor-VAE are shown in \figref{fig:vae_latent_traversal}. We experimented with $\beta$-VAEs but observed, that the reconstructions of are poor when we use higher $\beta$ in our settings. One potential reason is the discretely distributed positions in the trained datasets. The \position factor of \diagvib has nine possible values with large gap among them violating the Gaussian distribution assumption in the KL-divergence term during training. To empirically demonstrate this claim, we train $\beta$-VAEs with different $\beta$ on two settings: (1) with \position factor freely assigned to three different values (2) with \position factor fixed to the center of the images. According to the results shown in \figref{fig:vae_reconstruction_comparison}, the reconstructions from the training with fixed \position are significantly better, supporting the aforementioned argument.
%%% ZSO AND ZGO SUBSECTION
\subsection{ZSO and ZGO}
\label{supp:zso_zgo}
The mean accuracies \(\text{P}_i\) for all factors \(\mathcal{F}_i\) on the ZSO together with the mean accuracies \(\text{P}_{i,j}\) for all factor pairings \((\mathcal{F}_i,\mathcal{F}_j), i\neq j\) on the ZGO study for all baselines are shown in \figrefs{fig:cgo_rn18-dn,fig:cgo_wrn-vae}.
Additionally, the aggregated benchmark metrics \(\text{FAAvg}_i\) and \(\text{FAMin}_i\) for all factors \(\mathcal{F}_i, i\in\{0,1,...,6\}\) together with their respective standard errors for all baselines on the ZSO and ZGO studies are presented in \tabrefs{tab:zso,tab:zgo}.

\begin{figure*}[p!]
\begin{center}
	\includegraphics[width=0.8\linewidth]{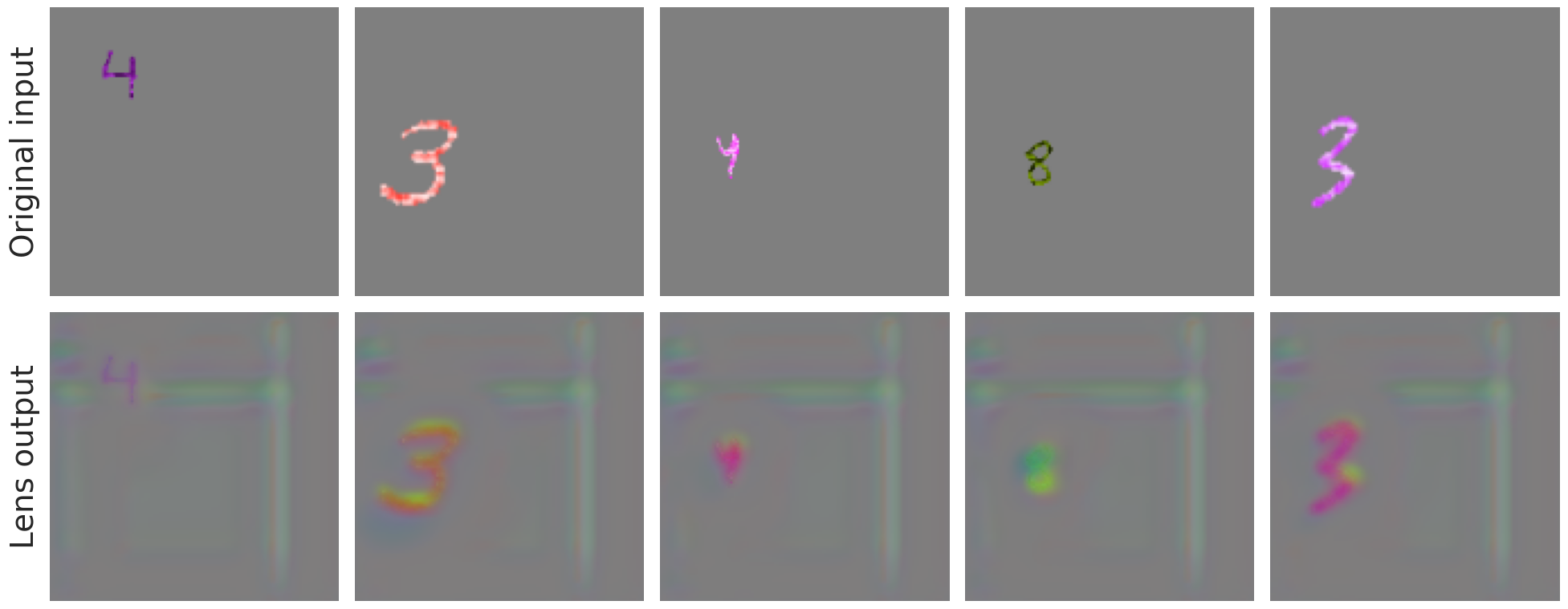}
	\caption{Examples of Automatic shortcut removal Lens output for ZSO study with factor \texture.}
	\label{fig:ASR_ZSO_texture_images}
\end{center}
\end{figure*}

From the exemplar VAE decoder reconstructions (\figref{fig:vae_reconstruction_comparison}) and ASR lens outputs (\figref{fig:ASR_ZSO_texture_images}) we find that the high-frequency texture information is not preserved, leading to the observed low classification accuracies on the texture task.
%%% CGO SUBSECTION
\subsection{Compositional-based generalization opportunities}
\label{supp:cgo}
\begin{figure*}[!htb]
	\begin{center}
		\includegraphics[width=\linewidth]{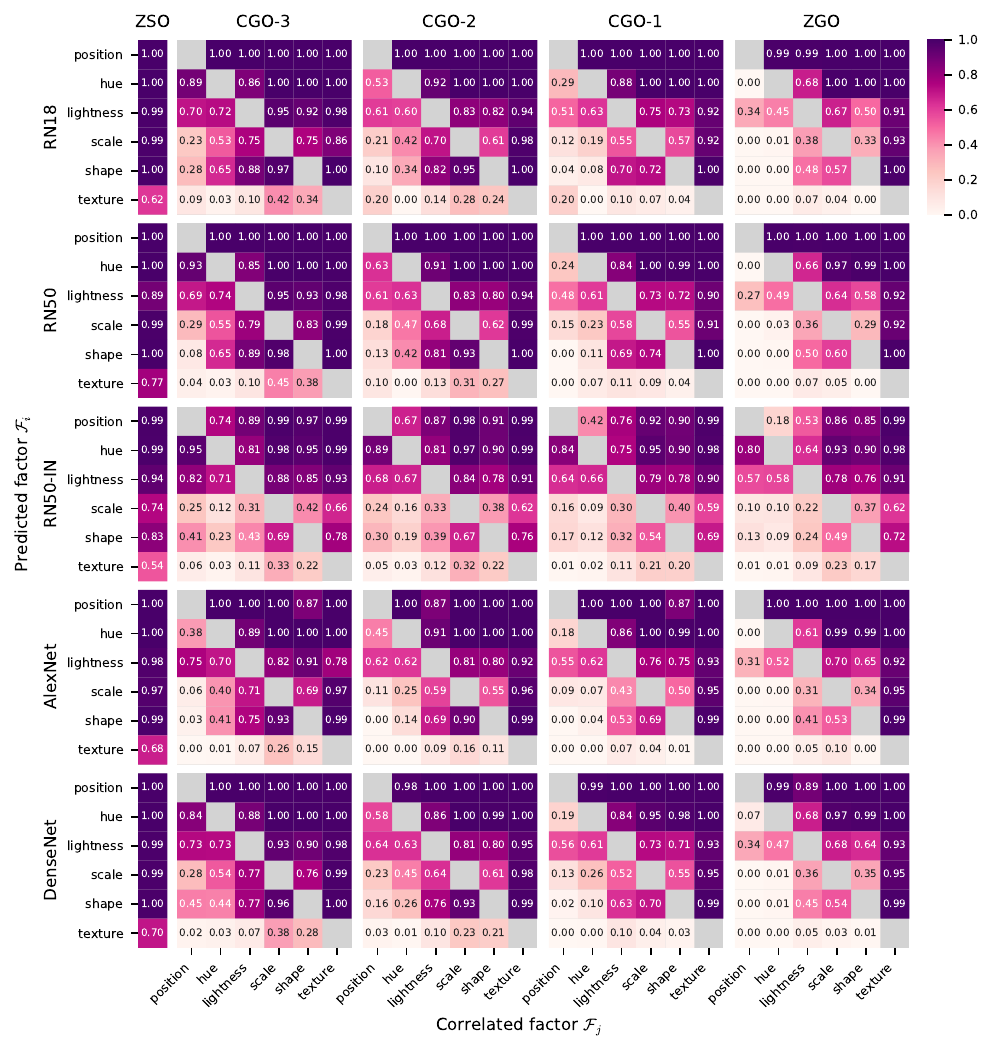}
		\caption{Mean accuracies \(\text{P}_i\) for all factors \(\mathcal{F}_i\) on the ZSO study and mean accuracies \(\text{P}_{i,j}\) for all factor pairings \((\mathcal{F}_i,\mathcal{F}_j), i\neq j\) for the RN18, RN50, RN50-IN, AlexNet and DenseNet on the CGO and ZGO studies.}
		\label{fig:cgo_rn18-dn}
	\end{center}
\end{figure*}

\begin{figure*}[!htb]
	\begin{center}
		\includegraphics[width=\linewidth]{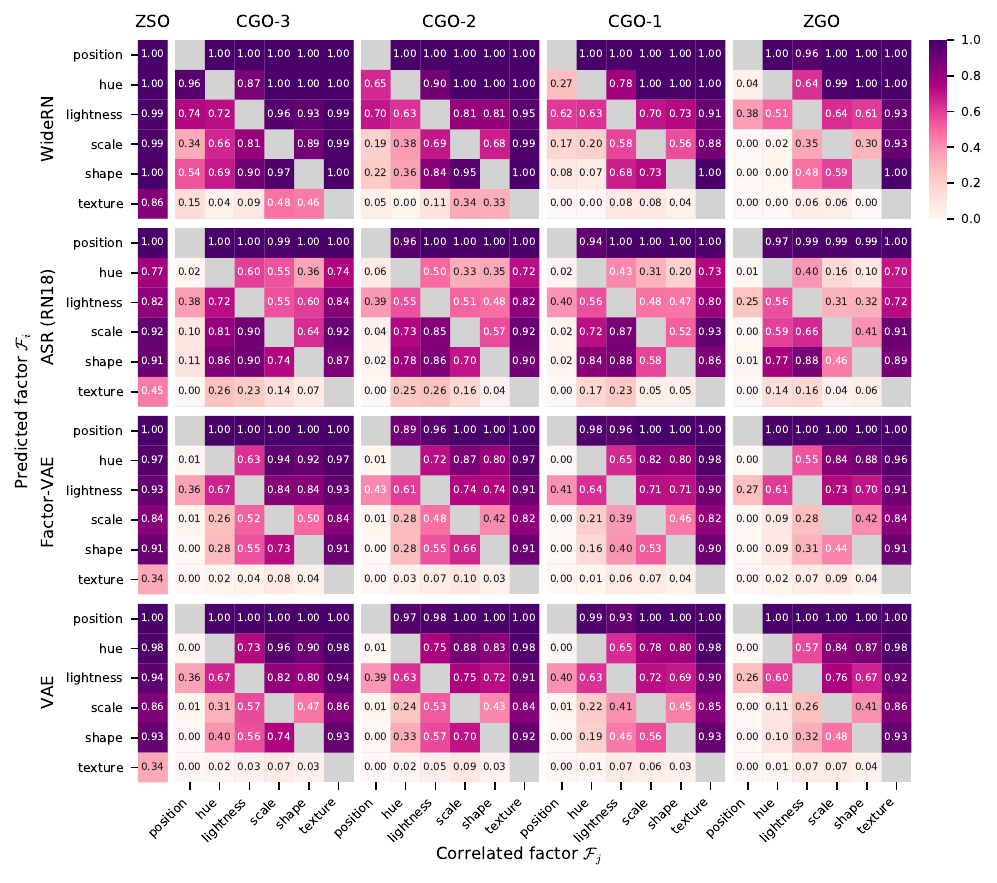}
		\caption{Mean accuracies \(\text{P}_i\) for all factors \(\mathcal{F}_i\) on the ZSO study and mean accuracies \(\text{P}_{i,j}\) for all factor pairings \((\mathcal{F}_i,\mathcal{F}_j), i\neq j\) for the WideRN, ASR (RN18), Factor-VAE and VAE on the CGO and ZGO studies.}
		\label{fig:cgo_wrn-vae}
	\end{center}
\end{figure*}

The mean accuracies \(\text{P}_i\) for all factors \(\mathcal{F}_i\) on the ZSO together with the mean accuracies \(\text{P}_{i,j}\) for all factor pairings \((\mathcal{F}_i,\mathcal{F}_j), i\neq j\) on the CGO and ZGO studies for all baselines are shown in \figrefs{fig:cgo_rn18-dn,fig:cgo_wrn-vae}.
Additionally, the aggregated benchmark metrics \(\text{FAAvg}_i\) and \(\text{FAMin}_i\) for all factors \(\mathcal{F}_i, i\in\{0,1,...,6\}\) together with their respective standard errors for all baselines on the CGO studies are presented in \tabrefs{tab:cgo_1,tab:cgo_2,tab:cgo_3}.

%%% FGO SUBSECTION
\subsection{Frequency-based generalization opportunities}
\label{supp:fgo}
\begin{figure*}[!htb]
	\begin{center}
		\includegraphics[width=\linewidth]{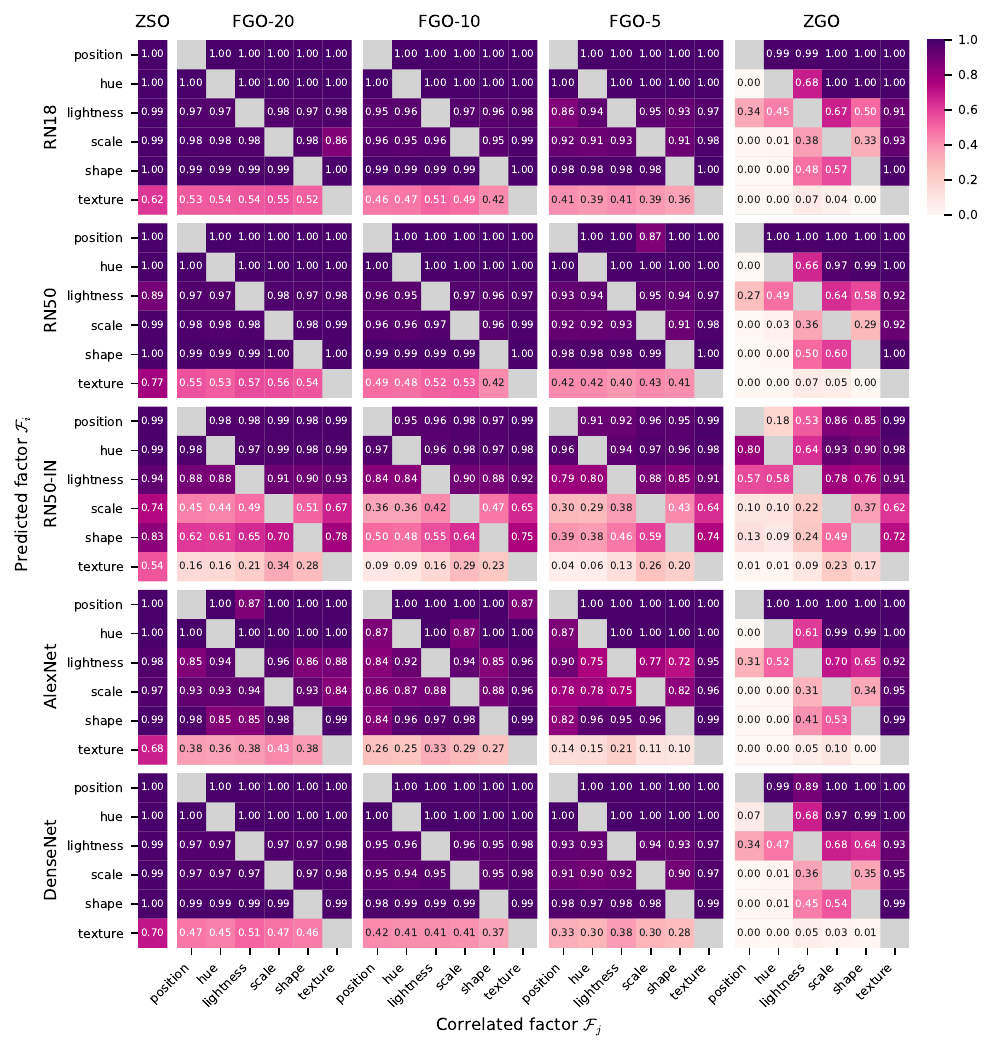}
		\caption{Mean accuracies \(\text{P}_i\) for all factors \(\mathcal{F}_i\) on the ZSO study and mean accuracies \(\text{P}_{i,j}\) for all factor pairings \((\mathcal{F}_i,\mathcal{F}_j), i\neq j\) for the RN18, RN50, RN50-IN, AlexNet and DenseNet on the FGO and ZGO studies.}		\label{fig:fgo_rn18-dn}
	\end{center}
\end{figure*}

\begin{figure*}[!htb]
	\begin{center}
		\includegraphics[width=\linewidth]{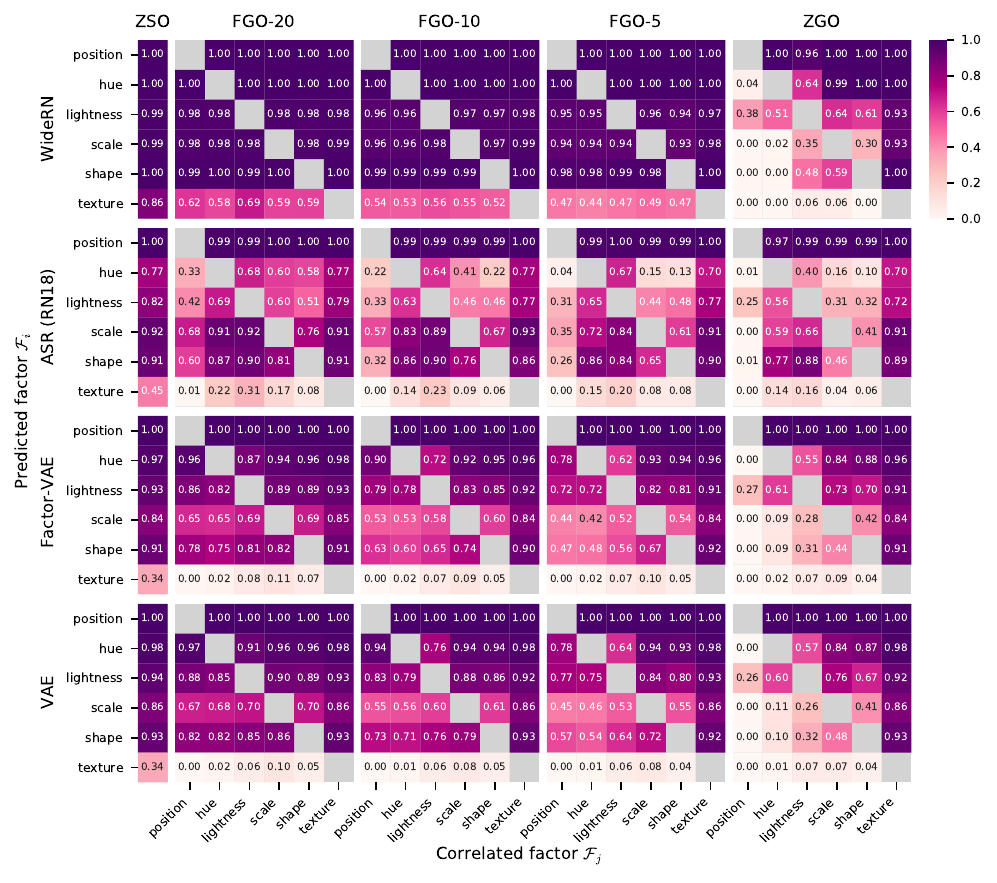}
		\caption{Mean accuracies \(\text{P}_i\) for all factors \(\mathcal{F}_i\) on the ZSO study and mean accuracies \(\text{P}_{i,j}\) for all factor pairings \((\mathcal{F}_i,\mathcal{F}_j), i\neq j\) for the WideRN, ASR (RN18), Factor-VAE and VAE on the FGO and ZGO studies.}
		\label{fig:fgo_wrn-vae}
	\end{center}
\end{figure*}

The mean accuracies \(\text{P}_i\) for all factors \(\mathcal{F}_i\) on the ZSO together with the mean accuracies \(\text{P}_{i,j}\) for all factor pairings \((\mathcal{F}_i,\mathcal{F}_j), i\neq j\) on the FGO and ZGO studies for all baselines are shown in \figrefs{fig:fgo_rn18-dn,fig:fgo_wrn-vae}.
Additionally, the aggregated benchmark metrics \(\text{FAAvg}_i\) and \(\text{FAMin}_i\) for all factors \(\mathcal{F}_i, i\in\{0,1,...,6\}\) together with their respective standard errors for all baselines on the FGO studies are presented in \tabrefs{tab:fgo_5,tab:fgo_10,tab:fgo_20}.

\begin{figure*}[!htb]
\begin{center}
	\includegraphics[width=0.8\linewidth]{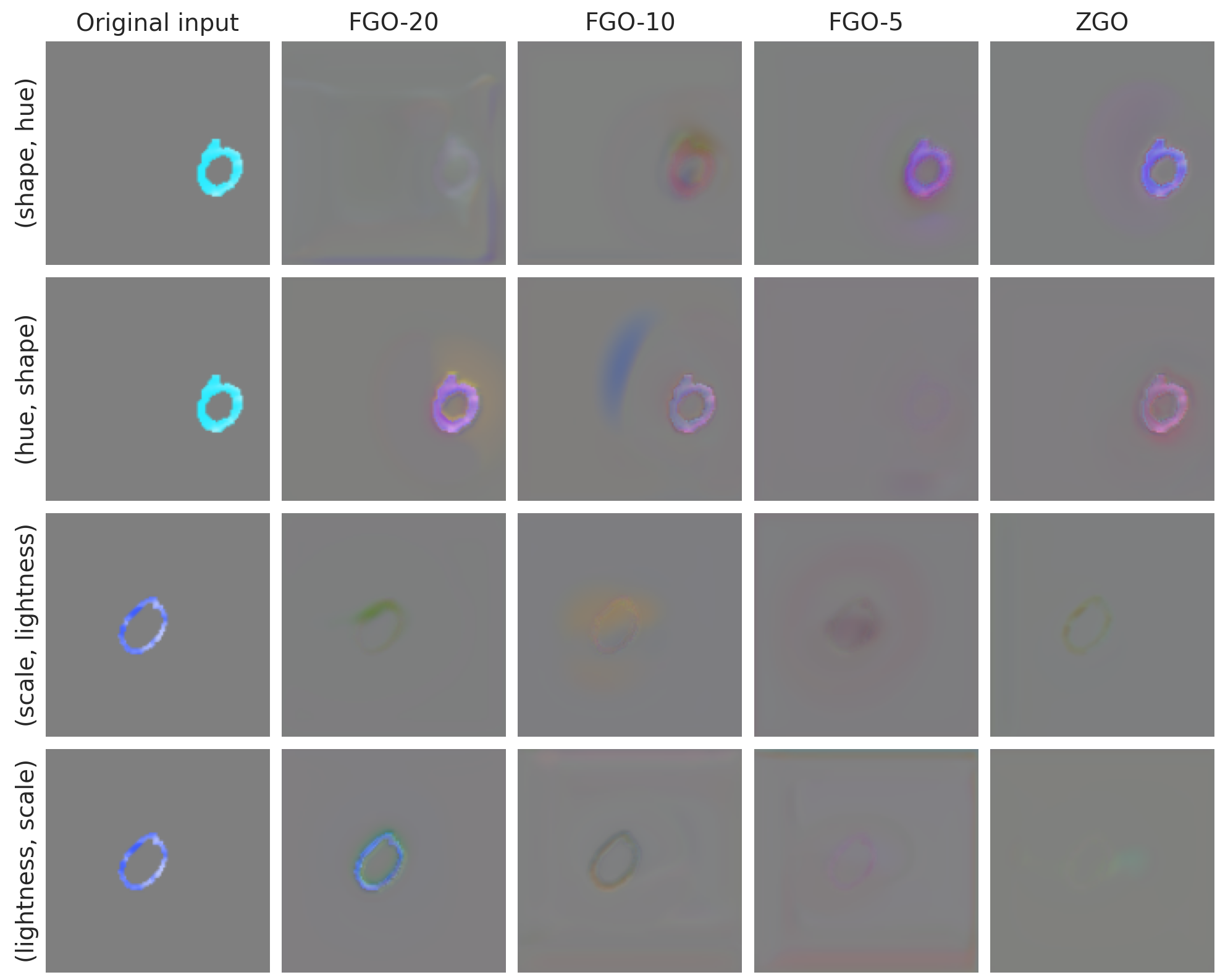}
	\caption{Examples of Automatic shortcut removal Lens output for different studies in our benchmark on datasets with following factor correlations: $(\shape, \hue)$ (first row), $(\hue, \shape)$ (second row), $(\scale, \lightness)$ (third row), $(\lightness, \scale)$ (fourth row).}
	\label{fig:ASR_images}
\end{center}
\end{figure*}

As discussed in \secref{sec:results}, we find that ASR fails to exploit the presented GO for \hue and \lightness. This is likely attributed to the pixel-wise reconstruction loss that is used as a regularizer in the ASR objective. We present qualitative results of this behaviour in \figref{fig:ASR_images}.

\input{figures/vae/figure_compared_traversal}
\input{figures/vae/figure_compared_center}

\begin{table*}[p!]
    \centering
    \setlength{\tabcolsep}{4.5pt}
    \input{tables/ZSO}
    \caption{FAAvg and respecive standard error for all baselines on the ZSO study.}
    \label{tab:zso}
\bigskip
\bigskip
    \centering
    \setlength{\tabcolsep}{4.5pt}
    \input{tables/FGO_100}
    \caption{FAAvg and FAMin together with their respective standard errors for all baselines on the ZGO study.}
    \label{tab:zgo}
\end{table*}


\begin{table*}[!h]
    \centering
    \setlength{\tabcolsep}{4.5pt}
    \input{tables/CGO_1}
    \caption{Aggregated benchmark metrics \(\text{FAAvg}_i\) and \(\text{FAMin}_i\) for all factors \(\mathcal{F}_i, i\in\{0,1,...,6\}\) together with their respective standard errors for all baselines on the CGO-1 study.}
    \label{tab:cgo_1}
\bigskip
\bigskip
    \centering
    \setlength{\tabcolsep}{4.5pt}
    \input{tables/CGO_2}
    \caption{Aggregated benchmark metrics \(\text{FAAvg}_i\) and \(\text{FAMin}_i\) for all factors \(\mathcal{F}_i, i\in\{0,1,...,6\}\) together with their respective standard errors for all baselines on the CGO-2 study.}
    \label{tab:cgo_2}
\bigskip
\bigskip
    \centering
    \setlength{\tabcolsep}{4.5pt}
    \input{tables/CGO_3}
    \caption{Aggregated benchmark metrics \(\text{FAAvg}_i\) and \(\text{FAMin}_i\) for all factors \(\mathcal{F}_i, i\in\{0,1,...,6\}\) together with their respective standard errors for all baselines on the CGO-3 study.}
    \label{tab:cgo_3}
\end{table*}


\begin{table*}[!h]
    \centering
    \setlength{\tabcolsep}{4.5pt}
    \input{tables/FGO_95}
    \caption{Aggregated benchmark metrics \(\text{FAAvg}_i\) and \(\text{FAMin}_i\) for all factors \(\mathcal{F}_i, i\in\{0,1,...,6\}\) together with their respective standard errors for all baselines on the FGO-5 study.}
    \label{tab:fgo_5}
\bigskip
\bigskip
    \centering
    \setlength{\tabcolsep}{4.5pt}
    \input{tables/FGO_90}
    \caption{Aggregated benchmark metrics \(\text{FAAvg}_i\) and \(\text{FAMin}_i\) for all factors \(\mathcal{F}_i, i\in\{0,1,...,6\}\) together with their respective standard errors for all baselines on the FGO-10 study.}
    \label{tab:fgo_10}
\bigskip
\bigskip
    \centering
    \setlength{\tabcolsep}{4.5pt}
    \input{tables/FGO_80}
    \caption{Aggregated benchmark metrics \(\text{FAAvg}_i\) and \(\text{FAMin}_i\) for all factors \(\mathcal{F}_i, i\in\{0,1,...,6\}\) together with their respective standard errors for all baselines on the FGO-20 study.}
    \label{tab:fgo_20}
\end{table*}

%% file: tables/all_factors_overview.tex
\begin{tabular}{ccccc}\toprule
\(\mathcal{F}_i\)			& \(\mathcal{S}_i\)				& \(\mathcal{N}_i\)			& \(\mathcal{C}_{i,j}\)			& \(\mathcal{S}_{i,j}\)\\\midrule
\multirow{9}{*}{position} 	& \multirow{9}{*}{\([0, 1]^2\)} & \multirow{9}{*}{9} 		& top-left						& \([1/7, 2/7]\times[1/7, 2/7]\)\\
						  															& & & top-center					& \([1/7, 2/7]\times[3/7, 4/7]\)\\
						  															& & & top-right						& \([1/7, 2/7]\times[5/7, 6/7]\)\\
						  															& & & center-left					& \([3/7, 4/7]\times[1/7, 2/7]\)\\
						  															& & & center-center					& \([3/7, 4/7]\times[3/7, 4/7]\)\\
						  															& & & center-right					& \([3/7, 4/7]\times[5/7, 6/7]\)\\
						  															& & & bottom-left					& \([5/7, 6/7]\times[1/7, 2/7]\)\\
						  															& & & bottom-center					& \([5/7, 6/7]\times[3/7, 4/7]\)\\
						  															& & & bottom-right					& \([5/7, 6/7]\times[5/7, 6/7]\)\\
\multirow{6}{*}{hue}			& \multirow{6}{*}{\([0, 2\pi)\)} & \multirow{6}{*}{6} 		& red 							& \([\ang{345}, \ang{15}]\)\\
																	 				& & & yellow						& \([\ang{45}, \ang{75}]\)\\
																	 				& & & green							& \([\ang{105}, \ang{135}]\)\\
																	 				& & & cyan							& \([\ang{165}, \ang{195}]\)\\
																	 				& & & blue							& \([\ang{225}, \ang{255}]\)\\
																	 				& & & magenta						& \([\ang{285}, \ang{315}]\)\\
\multirow{4}{*}{lightness}	& \multirow{4}{*}{\([0, 1]^2\)} & \multirow{4}{*}{4} 		& dark 							& \([0, 1/11]\times[4/11, 5/11]\)\\
																	 				& & & darker						& \([2/11, 3/11]\times[6/11, 7/11]\)\\
																	 				& & & brighter						& \([4/11, 5/11]\times[8/11, 9/11]\)\\
																	 				& & & bright						& \([6/11, 7/11]\times[10/11, 1.]\)\\
\multirow{5}{*}{scale}		& \multirow{5}{*}{\([1/1.45, 1.45]\)} & \multirow{5}{*}{5} 		& small							& \([1/1.45, 1/1.35]\)\\
																	 				& & & smaller						& \([1/1.25, 1/1.15]\)\\
																	 				& & & normal						& \([1/1.05, 1.05]\)\\
																	 				& & & larger						& \([1.15, 1.25]\)\\
																	 				& & & large							& \([1.35, 1.45]\)\\
\multirow{4}{*}{shape}	 	& \multirow{4}{*}{MNIST} 		& \multirow{4}{*}{\elias{10}} 		& `0'							& digits `0'\\
						  															& & & `1'							& digits `1'\\
						  															& & & \(\vdots\)					& \(\vdots\)\\
%    						  															& & & `2'							& digits `2'\\
%    						  															& & & `3'							& digits `3'\\
%    						  															& & & `4'							& digits `4'\\
%    						  															& & & `5'							& digits `5'\\
%    						  															& & & `6'							& digits `6'\\
%    						  															& & & `7'							& digits `7'\\
%    						  															& & & `8'							& digits `8'\\
%    						  															& & & `9'							& digits `9'\\
\multirow{5}{*}{texture}	 	& \multirow{5}{*}{textures} & \multirow{5}{*}{5} 		& tiles								& tiles texture crops\\
						  															& & & wood							& wood texture crops\\
						  															& & & carpet						& carpet texture crops\\
						  															& & & bricks						& bricks texture crops\\
						  															& & & lava							& lava texture crops\\
\end{tabular}

%% file: figures/vae/figure_compared_traversal.tex
\begin{figure*}[p!]
    \centering
    \begin{subfigure}[b]{0.45\textwidth}
        \centering
        \includegraphics[width=\textwidth]{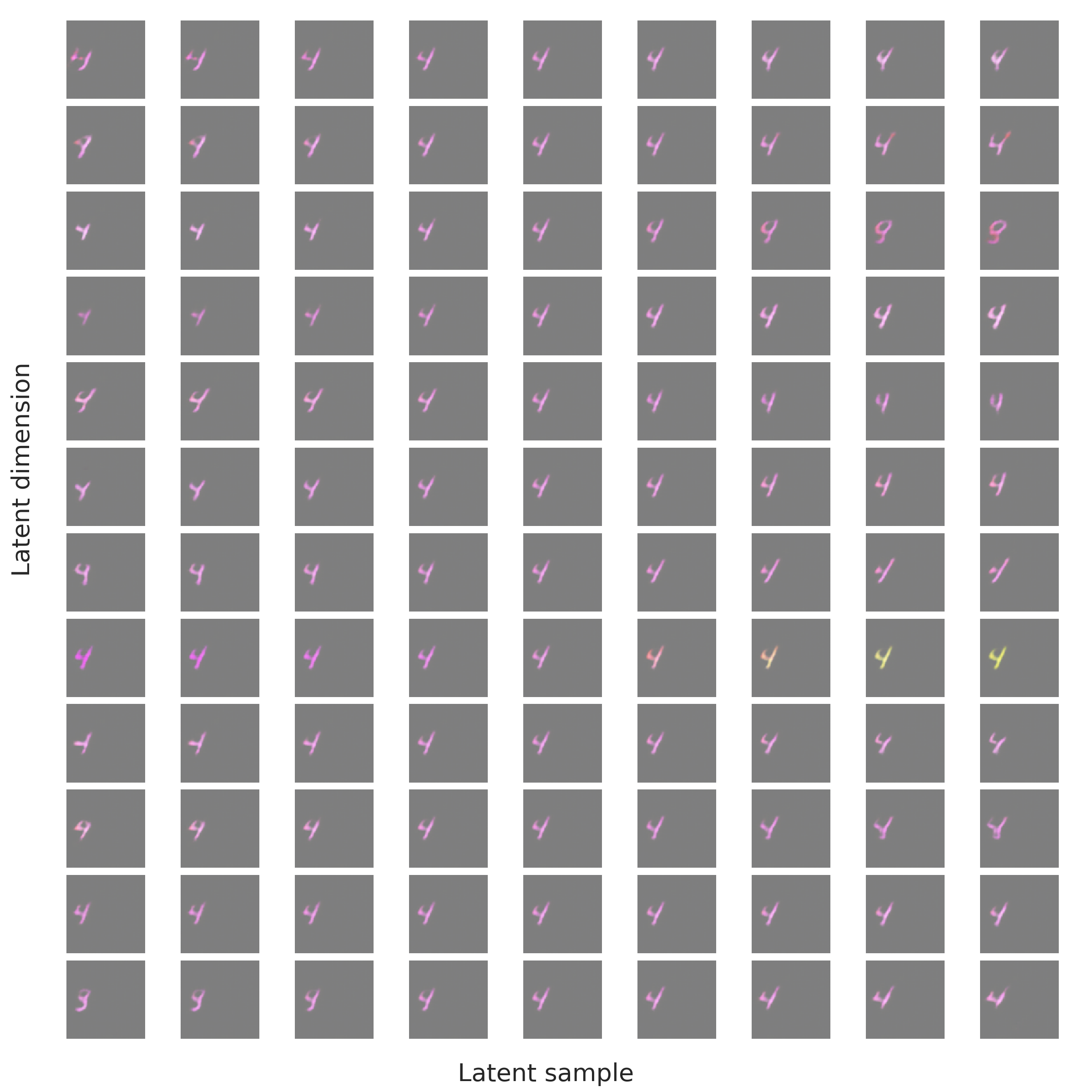}
        \caption{Latent traversal plot from VAE}
    \end{subfigure}
    \begin{subfigure}[b]{0.45\textwidth}
        \centering
        \includegraphics[width=\textwidth]{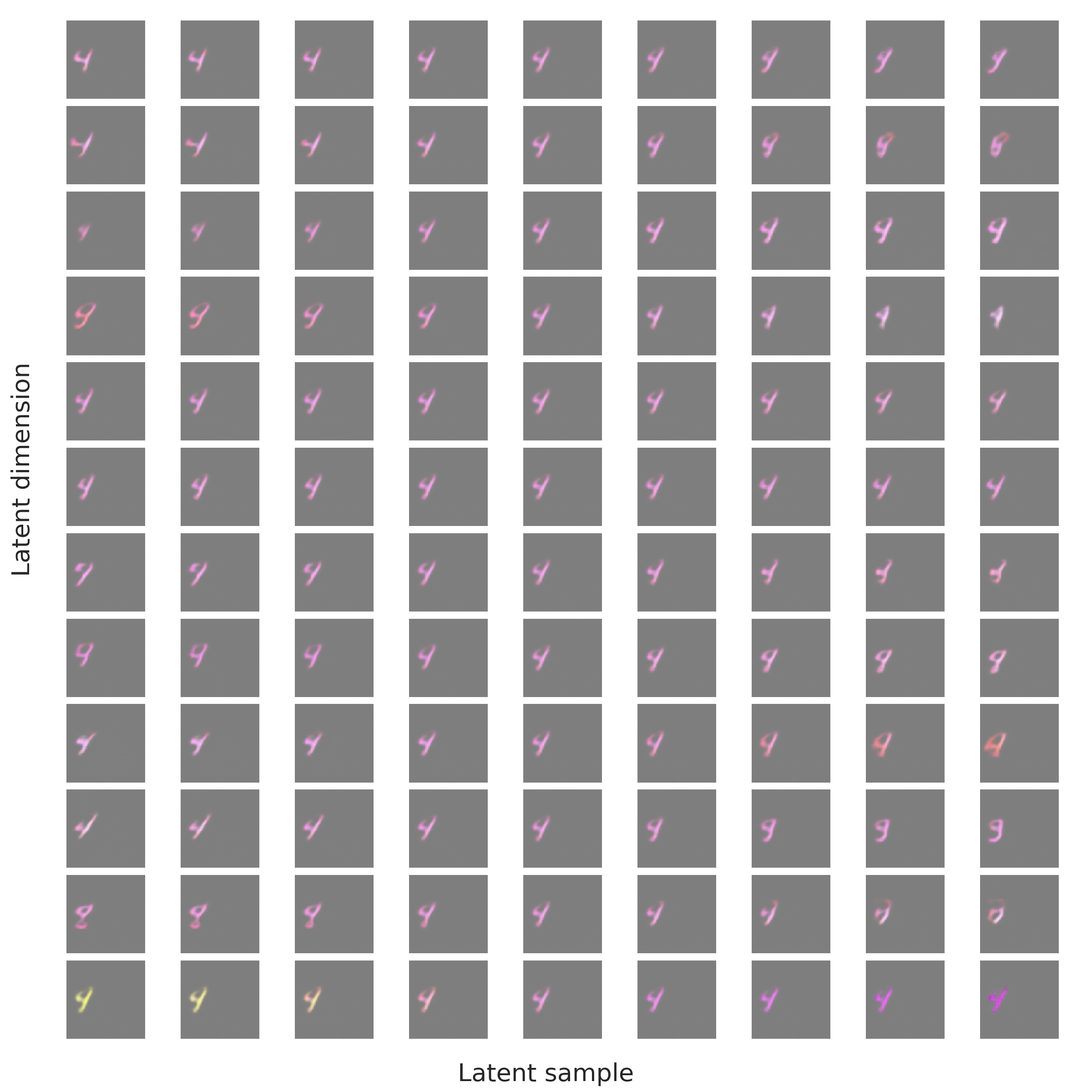}
        \caption{Latent traversal plot from Factor-VAE}
    \end{subfigure}
    \hfill
    \begin{subfigure}[b]{0.45\textwidth}
        \centering
        \includegraphics[width=\textwidth]{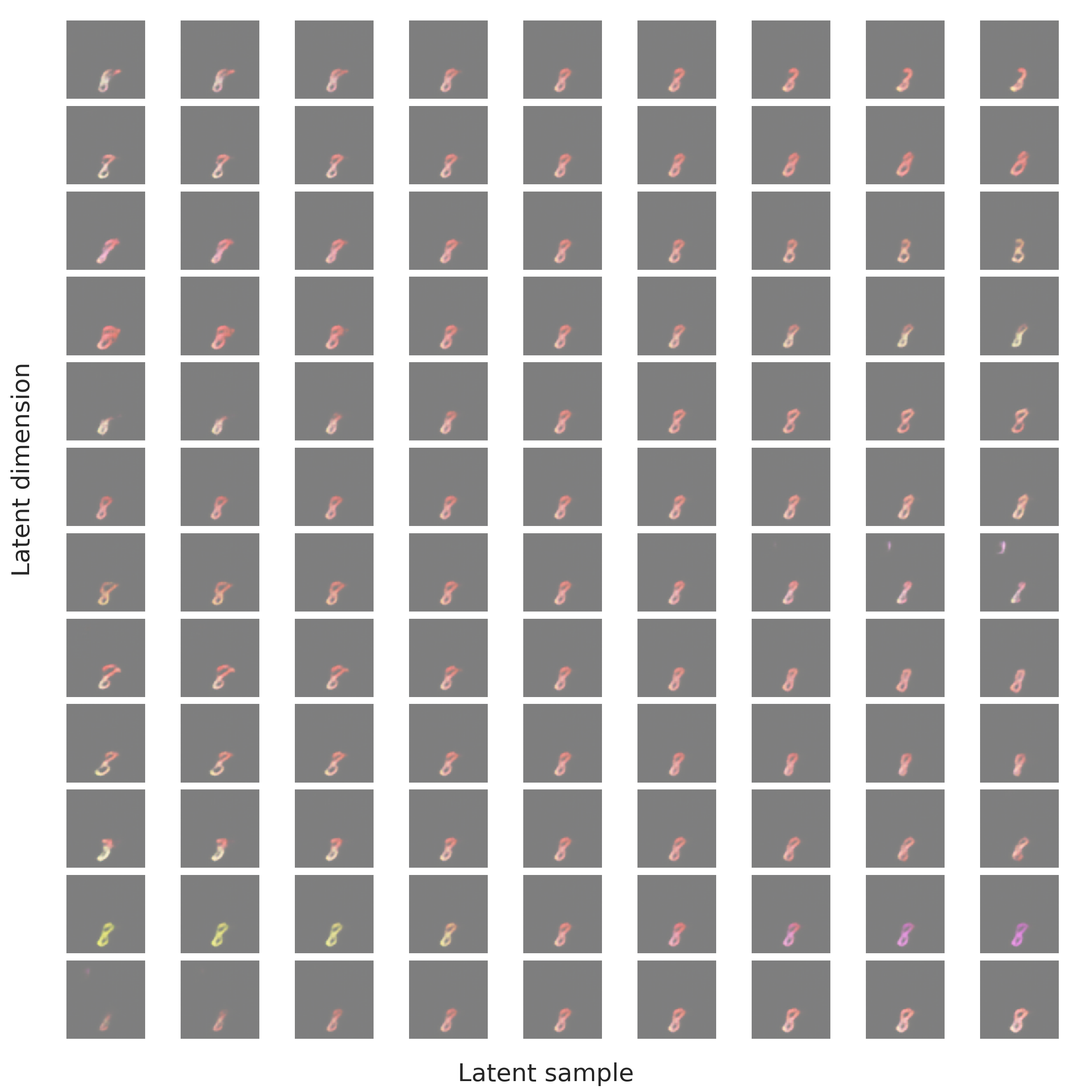}
        \caption{Latent traversal plot from VAE}
    \end{subfigure}
    \begin{subfigure}[b]{0.45\textwidth}
        \centering
        \includegraphics[width=\textwidth]{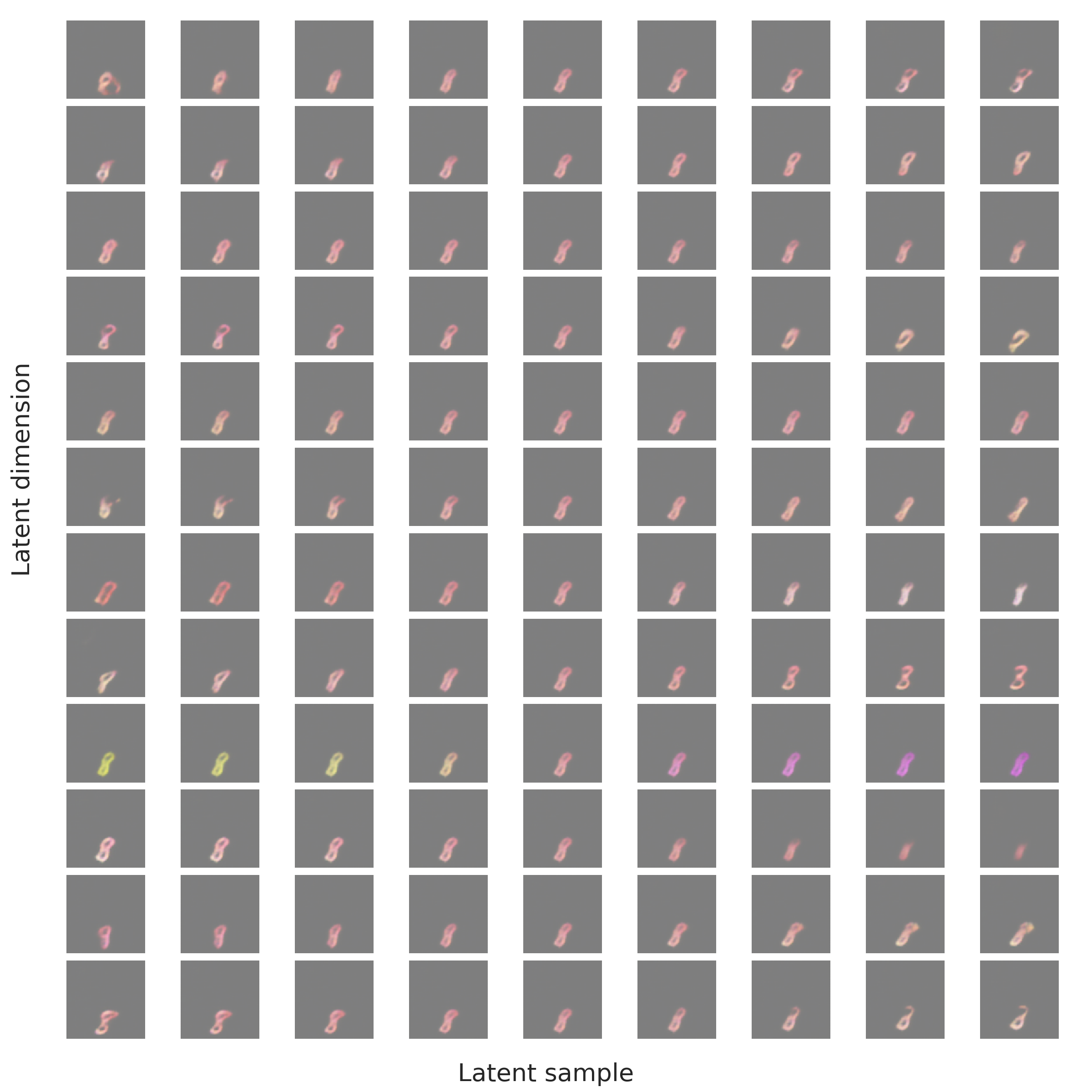}
        \caption{Latent traversal plot from Factor-VAE}
    \end{subfigure}     
    \caption{Comparison of latent traversals from VAE and Factor-VAE}
    \label{fig:vae_latent_traversal}
\end{figure*}

%% file: figures/vae/figure_compared_center.tex
\begin{figure}[p!]
\vspace{-\baselineskip}
     \centering
     \begin{subfigure}{0.23\textwidth}
         \centering
         \includegraphics[width=\textwidth]{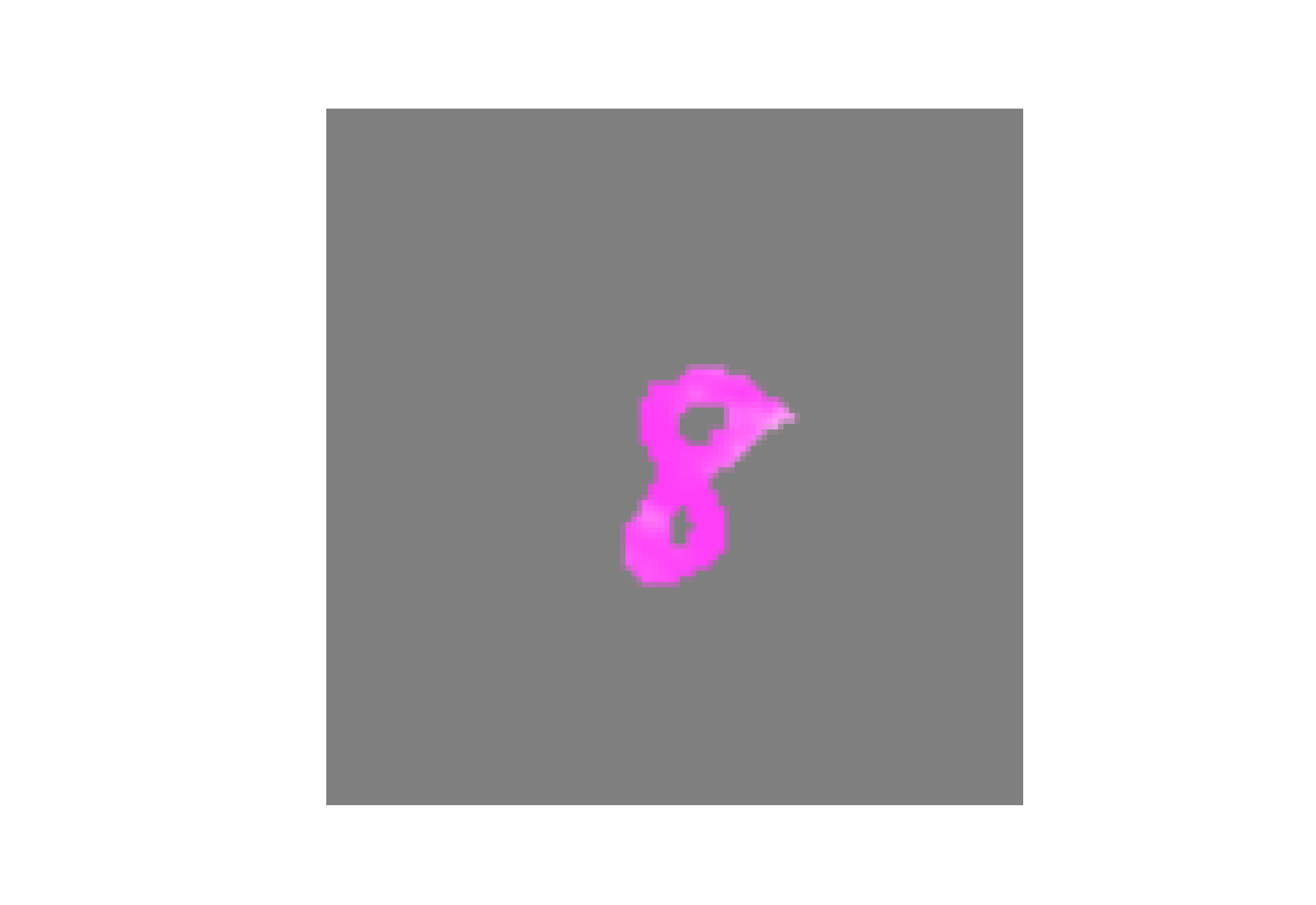}
         \captionsetup{justification=centering}
         \caption{Input image (Sample 1)}
     \end{subfigure}
     \begin{subfigure}{0.23\textwidth}
         \centering
         \includegraphics[width=\textwidth]{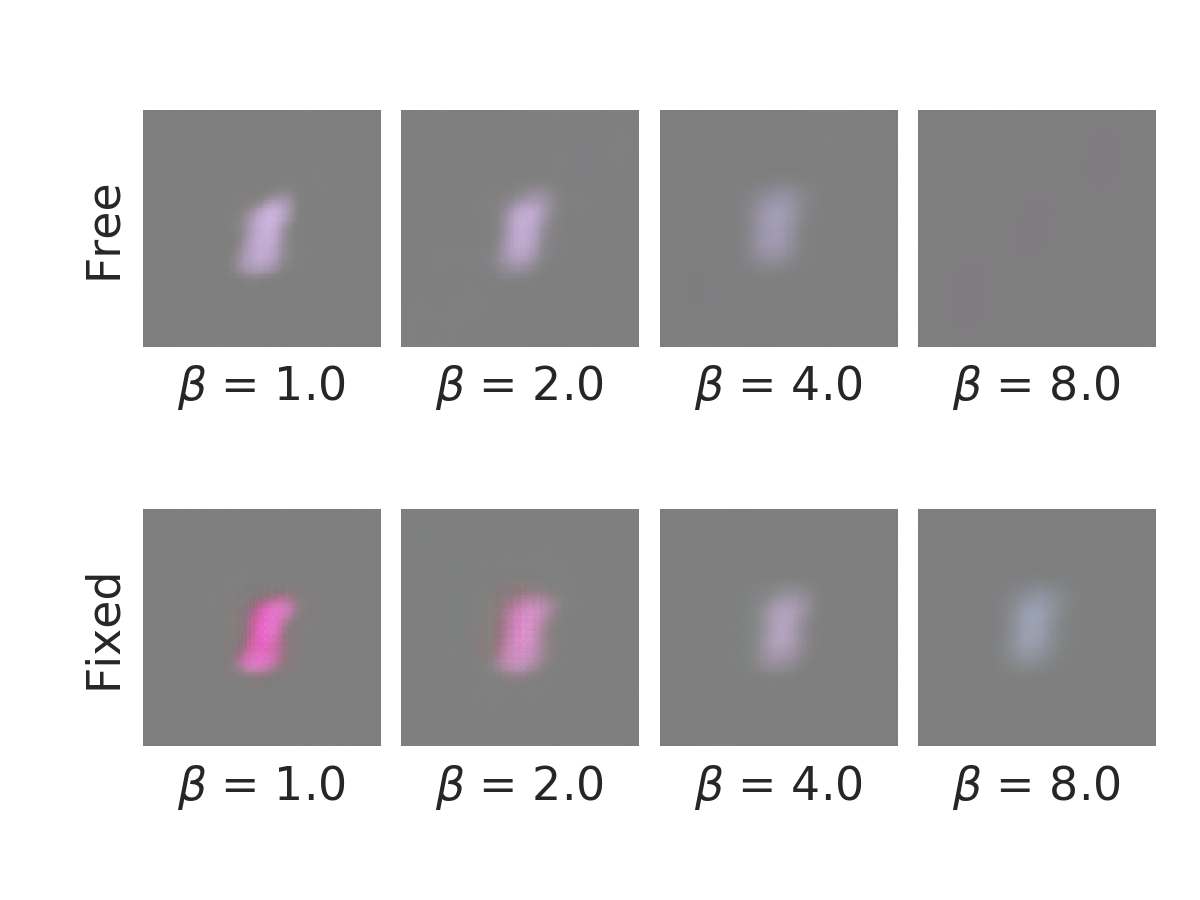}
         \captionsetup{justification=centering}
         \caption{Reconstructed images \newline (Sample 1)}
     \end{subfigure}
     \begin{subfigure}{0.23\textwidth}
         \centering
         \includegraphics[width=\textwidth]{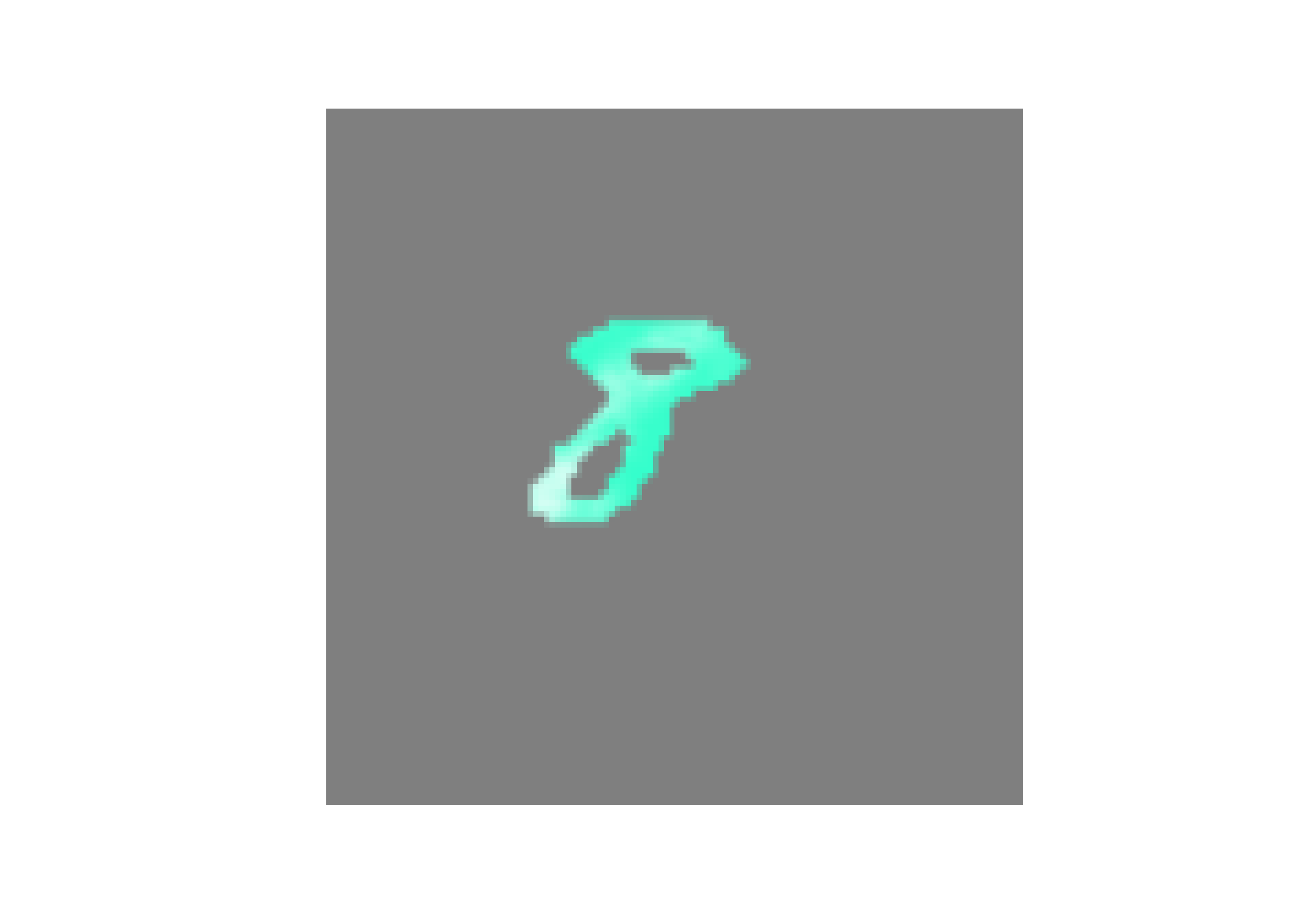}
         \captionsetup{justification=centering}
         \caption{Input image (Sample 2)}
     \end{subfigure}
     \begin{subfigure}{0.23\textwidth}
         \centering
         \includegraphics[width=\textwidth]{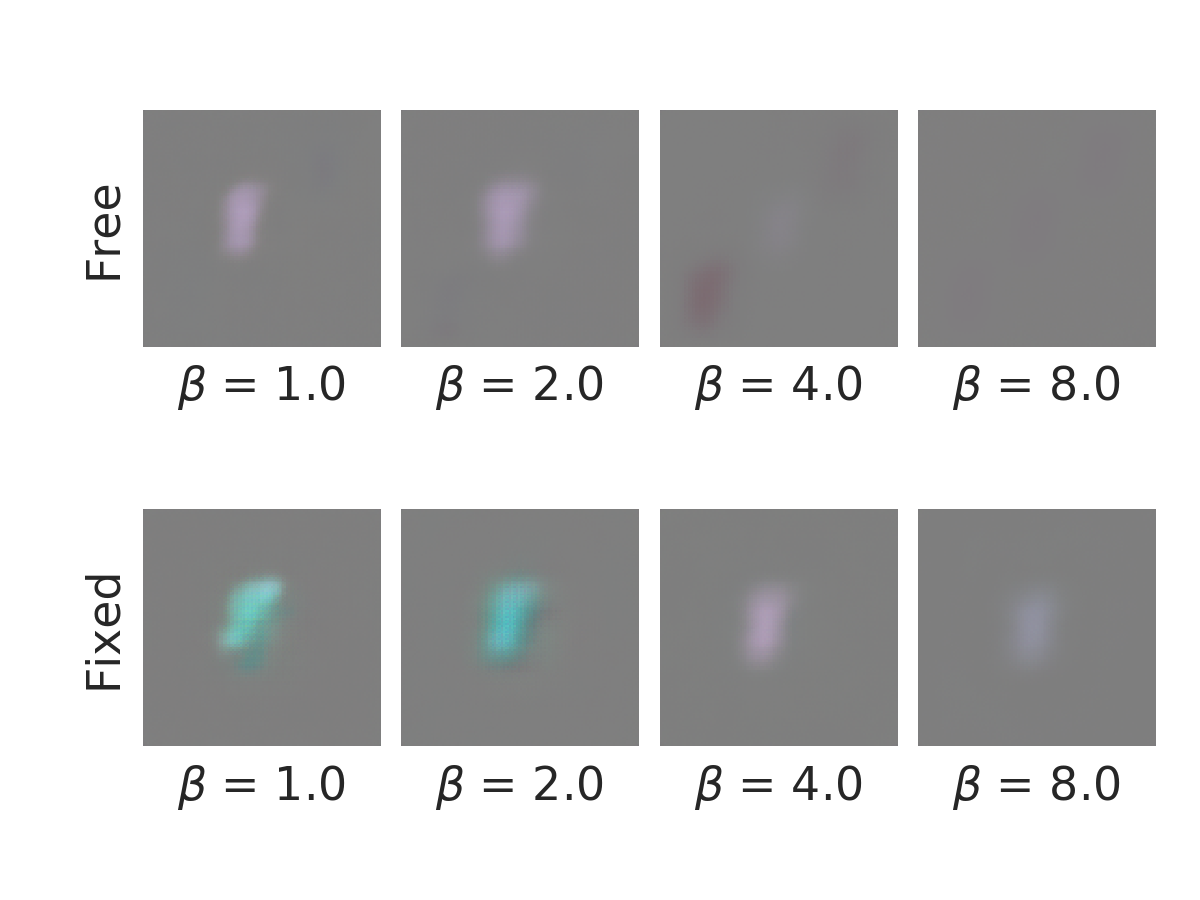}
         \captionsetup{justification=centering}
         \caption{Reconstructed images \newline (Sample 2)}
     \end{subfigure}     
     \hfill
     \begin{subfigure}{0.23\textwidth}
         \centering
         \includegraphics[width=\textwidth]{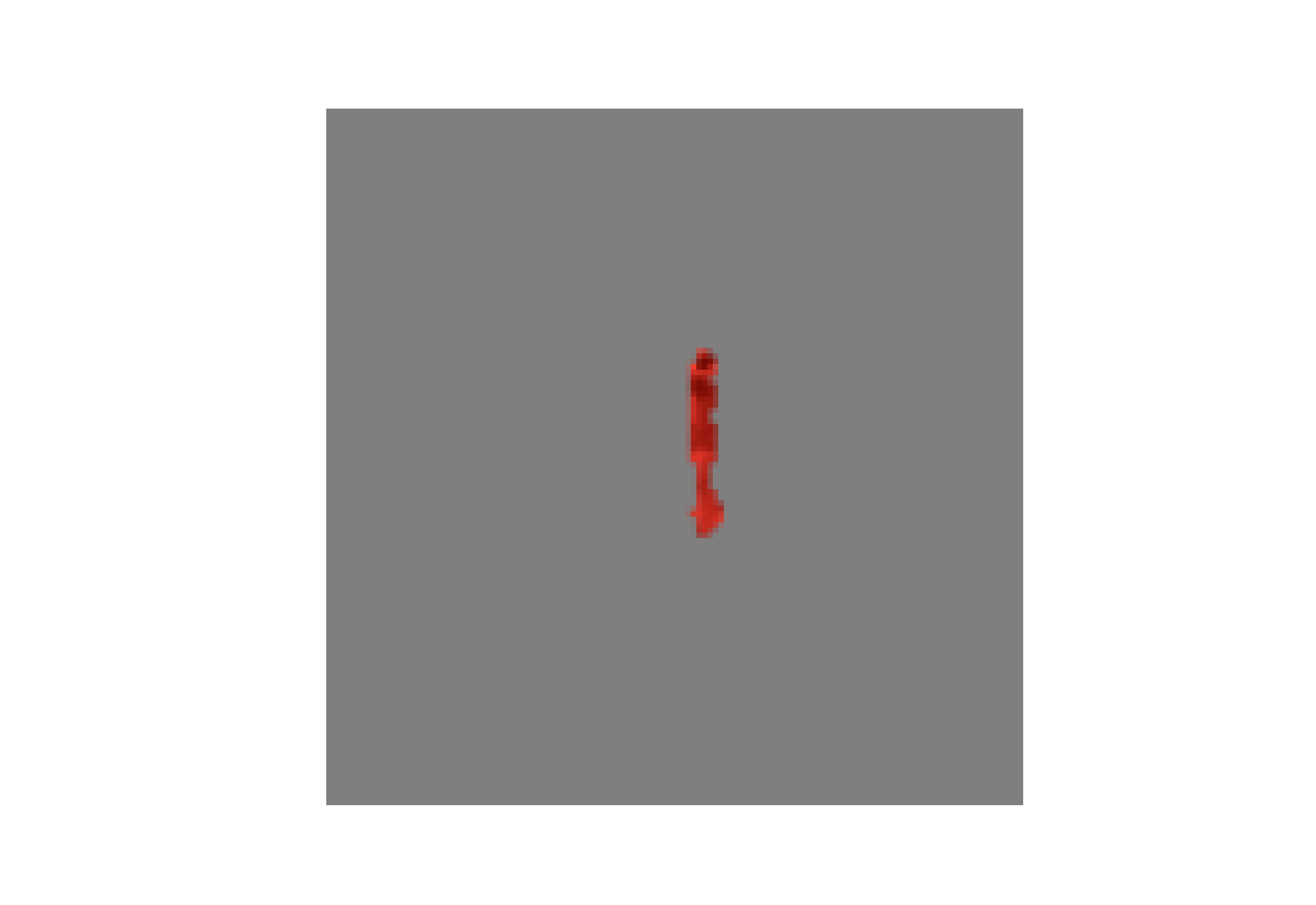}
         \captionsetup{justification=centering}
         \caption{Input image (Sample 3)}
     \end{subfigure}
     \begin{subfigure}{0.23\textwidth}
         \centering
         \includegraphics[width=\textwidth]{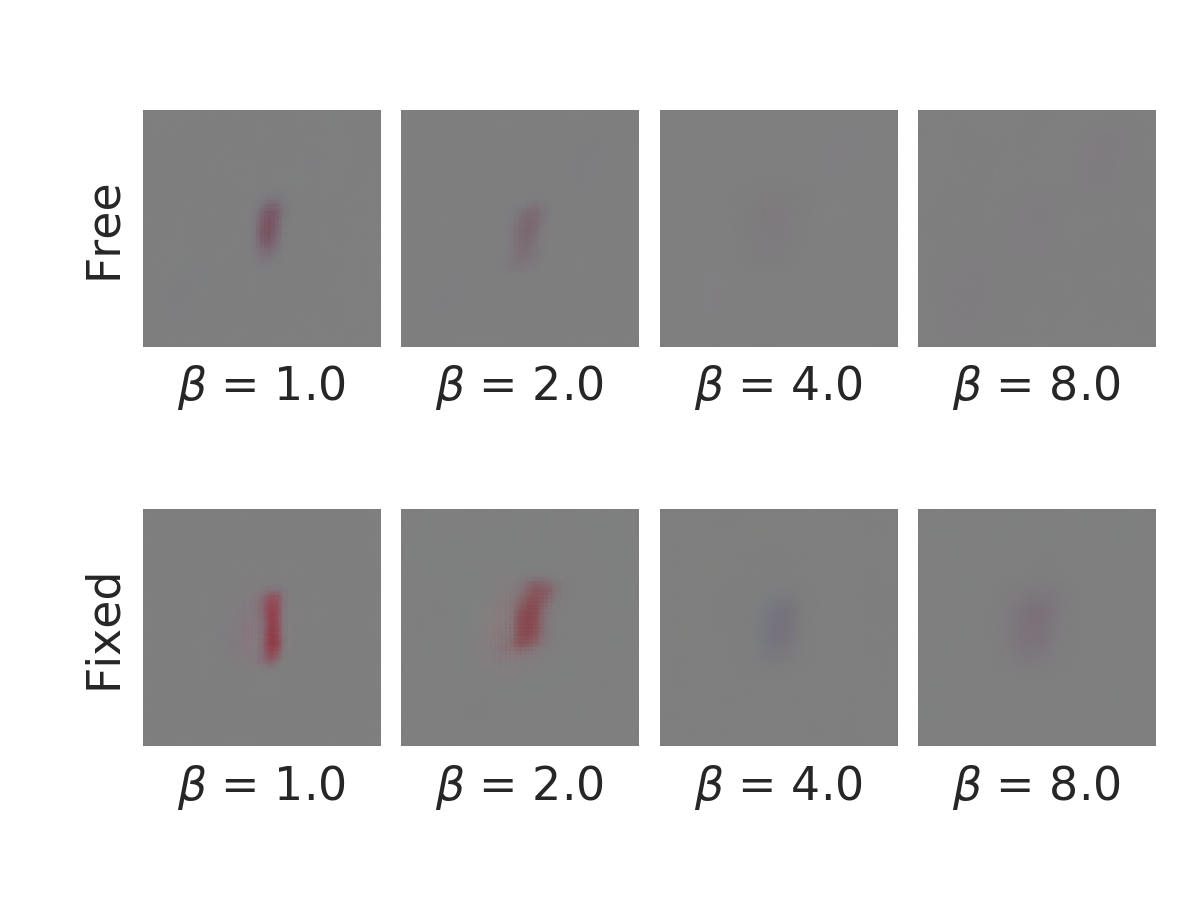}
         \captionsetup{justification=centering}
         \caption{Reconstructed images \newline (Sample 3)}
     \end{subfigure}
     \begin{subfigure}{0.23\textwidth}
         \centering
         \includegraphics[width=\textwidth]{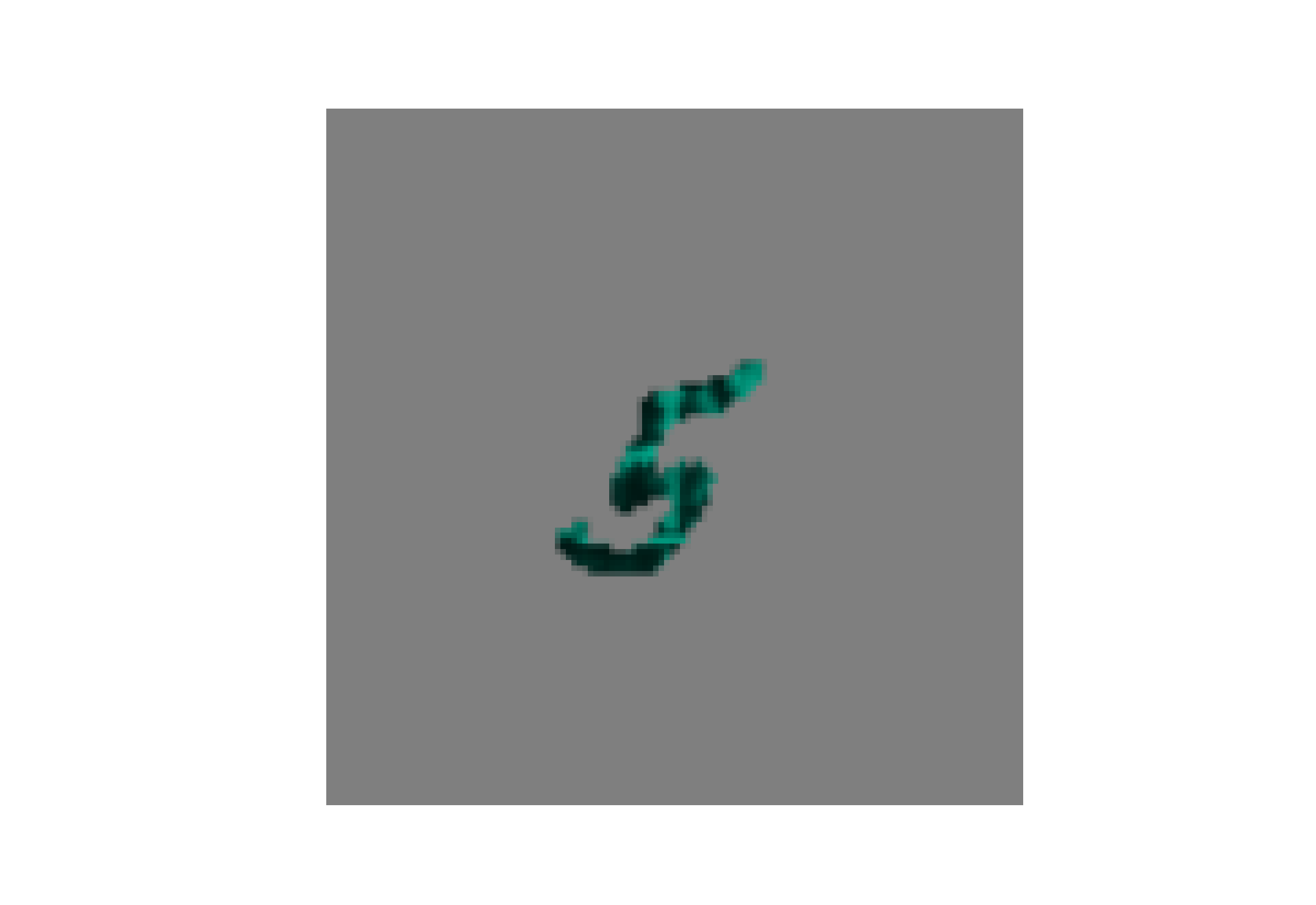}
         \captionsetup{justification=centering}
         \caption{Input image (Sample 4)}
     \end{subfigure}
     \begin{subfigure}{0.23\textwidth}
         \centering
         \includegraphics[width=\textwidth]{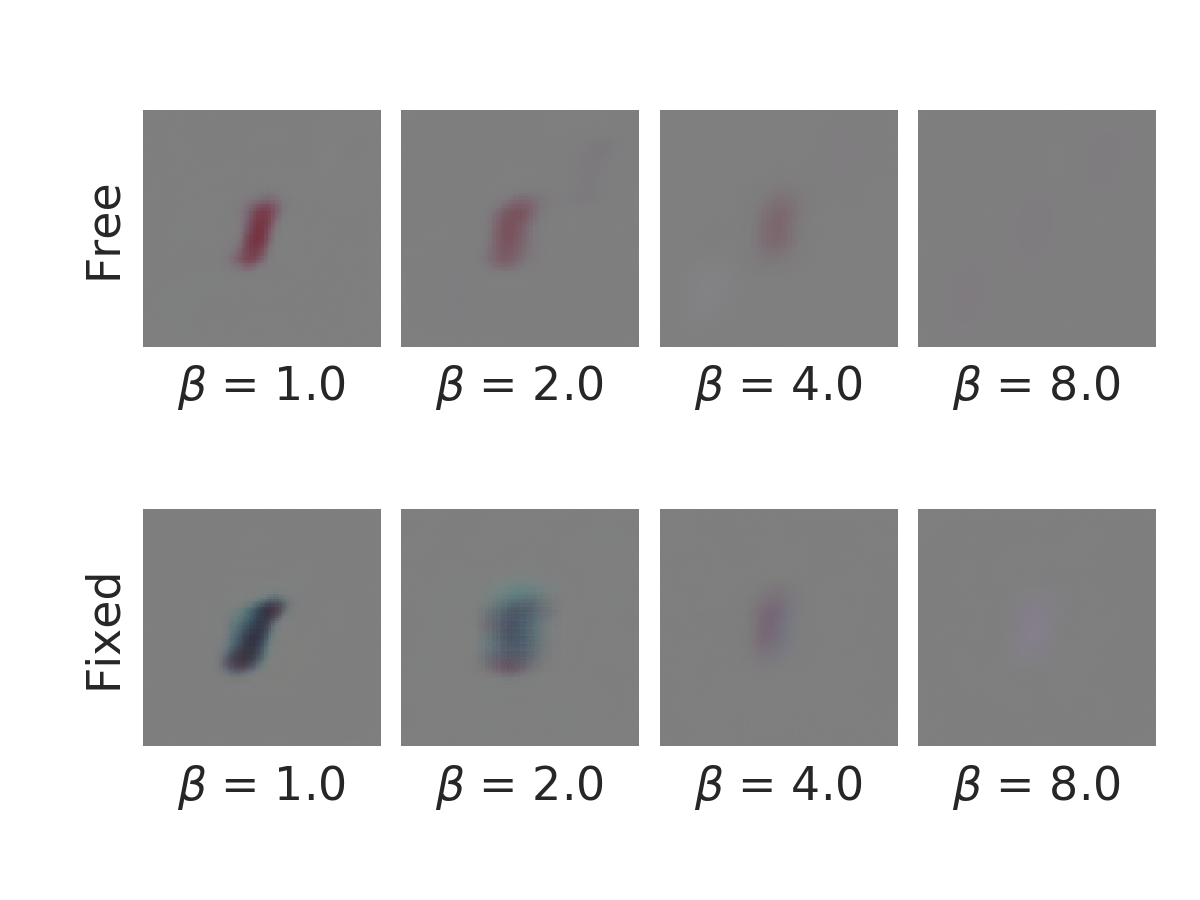}
         \captionsetup{justification=centering}
         \caption{Reconstructed images \newline (Sample 4)}
     \end{subfigure}     
    \caption{Reconstruction images from the $\beta$-VAE networks trained with datasets of free and fixed positions from different $\beta$ values}
    \label{fig:vae_reconstruction_comparison}
\vspace{-\baselineskip}
\end{figure}

%% file: tables/ZSO.tex
% ==================================================================== 
% This is an automatically generated tex file. 
% To update, please see: ./figures_and_tables/gen_table_ZSO.py 
% ==================================================================== 
\begin{tabular}{lS[table-format=2(1)]S[table-format=2(1)]S[table-format=2(1)]S[table-format=2(1)]S[table-format=2(1)]S[table-format=2(1)]} 
    \toprule 
     & {position}    & {hue}         & {lightness}   & {scale}       & {shape}       & {texture}     \\ 
    \midrule 
            RN18 & 100(0)    & 100(0)    & 99(0)     & 99(0)     & 100(0)    & 62(6)     \\ 
            RN50 & 100(0)    & 100(0)    & 89(9)     & 99(0)     & 100(0)    & 77(3)     \\ 
         RN50-IN & 99(0)     & 99(0)     & 94(1)     & 74(2)     & 83(3)     & 54(1)     \\ 
         AlexNet & 100(0)    & 100(0)    & 98(1)     & 97(0)     & 99(0)     & 68(3)     \\ 
        DenseNet & 100(0)    & 100(0)    & 99(0)     & 99(0)     & 100(0)    & 70(3)     \\ 
             WRN & 100(0)    & 100(0)    & 99(0)     & 99(0)     & 100(0)    & 86(2)     \\ 
      ASR (RN18) & 100(0)    & 77(6)     & 82(2)     & 92(1)     & 91(1)     & 45(1)     \\ 
      Factor-VAE & 100(0)    & 97(1)     & 93(2)     & 84(1)     & 91(2)     & 34(0)     \\ 
             VAE & 100(0)    & 98(1)     & 94(1)     & 86(2)     & 93(2)     & 34(0)     \\ 
    \bottomrule 
\end{tabular} 

%% file: tables/FGO_100.tex
% ==================================================================== 
% This is an automatically generated tex file. 
% To update, please see: ./figures_and_tables/gen_table_FGO.py 
% ==================================================================== 
\begin{tabular}{lS[table-format=2(1)]S[table-format=2(1)]S[table-format=2(1)]S[table-format=2(1)]S[table-format=2(1)]S[table-format=2(1)]S[table-format=2(1)]S[table-format=2(1)]S[table-format=2(1)]S[table-format=2(1)]S[table-format=2(1)]S[table-format=2(1)]} 
    \toprule 
     & \multicolumn{2}{c}{position} & \multicolumn{2}{c}{hue} & \multicolumn{2}{c}{lightness} & \multicolumn{2}{c}{scale} & \multicolumn{2}{c}{shape} & \multicolumn{2}{c}{texture} \\ 
    \cmidrule(lr){2-3}\cmidrule(lr){4-5}\cmidrule(lr){6-7}\cmidrule(lr){8-9}\cmidrule(lr){10-11}\cmidrule(lr){12-13} 
                 & {FAAvg} & {FAMin} & {FAAvg} & {FAMin} & {FAAvg} & {FAMin} & {FAAvg} & {FAMin} & {FAAvg} & {FAMin} & {FAAvg} & {FAMin} \\ 
    \midrule 
            RN18 & 100(0)    & 99(1)     & 74(2)     & 0(0)      & 57(4)     & 31(5)     & 33(2)     & 0(0)      & 41(2)     & 0(0)      & 2(1)      & 0(0)      \\ 
            RN50 & 100(0)    & 100(0)    & 72(2)     & 0(0)      & 58(4)     & 27(1)     & 32(2)     & 0(0)      & 42(2)     & 0(0)      & 3(1)      & 0(0)      \\ 
         RN50-IN & 68(3)     & 18(2)     & 85(2)     & 64(3)     & 72(4)     & 50(6)     & 28(2)     & 4(1)      & 33(5)     & 8(4)      & 10(1)     & 1(0)      \\ 
         AlexNet & 100(0)    & 100(0)    & 72(2)     & 0(0)      & 62(6)     & 31(5)     & 32(2)     & 0(0)      & 39(2)     & 0(0)      & 3(1)      & 0(0)      \\ 
        DenseNet & 98(1)     & 88(6)     & 74(3)     & 7(3)      & 61(6)     & 34(8)     & 34(1)     & 0(0)      & 40(2)     & 0(0)      & 2(0)      & 0(0)      \\ 
             WRN & 99(1)     & 96(3)     & 73(3)     & 4(3)      & 61(6)     & 37(8)     & 32(2)     & 0(0)      & 41(2)     & 0(0)      & 2(1)      & 0(0)      \\ 
      ASR (RN18) & 99(0)     & 96(1)     & 27(3)     & 1(0)      & 43(2)     & 25(0)     & 51(2)     & 0(0)      & 60(1)     & 1(0)      & 8(0)      & 0(0)      \\ 
      Factor-VAE & 100(0)    & 100(0)    & 65(4)     & 0(0)      & 64(4)     & 27(0)     & 32(1)     & 0(0)      & 35(2)     & 0(0)      & 4(0)      & 0(0)      \\ 
             VAE & 100(0)    & 100(0)    & 65(4)     & 0(0)      & 64(4)     & 26(0)     & 33(2)     & 0(0)      & 37(3)     & 0(0)      & 4(0)      & 0(0)      \\ 
    \bottomrule 
\end{tabular} 

%% file: tables/CGO_1.tex
% ==================================================================== 
% This is an automatically generated tex file. 
% To update, please see: ./figures_and_tables/gen_table_CGO.py 
% ==================================================================== 
\begin{tabular}{lS[table-format=2(1)]S[table-format=2(1)]S[table-format=2(1)]S[table-format=2(1)]S[table-format=2(1)]S[table-format=2(1)]S[table-format=2(1)]S[table-format=2(1)]S[table-format=2(1)]S[table-format=2(1)]S[table-format=2(1)]S[table-format=2(1)]} 
    \toprule 
     & \multicolumn{2}{c}{position} & \multicolumn{2}{c}{hue} & \multicolumn{2}{c}{lightness} & \multicolumn{2}{c}{scale} & \multicolumn{2}{c}{shape} & \multicolumn{2}{c}{texture} \\ 
    \cmidrule(lr){2-3}\cmidrule(lr){4-5}\cmidrule(lr){6-7}\cmidrule(lr){8-9}\cmidrule(lr){10-11}\cmidrule(lr){12-13} 
                 & {FAAvg} & {FAMin} & {FAAvg} & {FAMin} & {FAAvg} & {FAMin} & {FAAvg} & {FAMin} & {FAAvg} & {FAMin} & {FAAvg} & {FAMin} \\ 
    \midrule 
            RN18 & 100(0)    & 100(0)    & 83(1)     & 29(5)     & 71(8)     & 49(9)     & 47(3)     & 6(2)      & 51(3)     & 2(1)      & 8(2)      & 0(0)      \\ 
            RN50 & 100(0)    & 100(0)    & 82(1)     & 24(6)     & 69(8)     & 46(8)     & 48(4)     & 9(3)      & 51(3)     & 0(0)      & 6(2)      & 0(0)      \\ 
         RN50-IN & 80(2)     & 42(4)     & 88(2)     & 73(3)     & 75(5)     & 59(8)     & 31(3)     & 8(2)      & 37(6)     & 12(5)     & 11(1)     & 1(0)      \\ 
         AlexNet & 97(2)     & 87(12)    & 81(2)     & 18(7)     & 72(8)     & 52(10)    & 41(3)     & 2(1)      & 45(3)     & 0(0)      & 3(0)      & 0(0)      \\ 
        DenseNet & 100(0)    & 99(1)     & 79(1)     & 19(1)     & 71(8)     & 50(10)    & 48(4)     & 9(3)      & 49(2)     & 2(1)      & 3(0)      & 0(0)      \\ 
             WRN & 100(0)    & 100(0)    & 81(2)     & 27(4)     & 72(9)     & 51(10)    & 48(3)     & 10(3)     & 51(3)     & 4(2)      & 4(1)      & 0(0)      \\ 
      ASR (RN18) & 99(1)     & 94(3)     & 34(3)     & 2(1)      & 54(4)     & 40(5)     & 61(2)     & 2(1)      & 64(3)     & 2(1)      & 10(1)     & 0(0)      \\ 
      Factor-VAE & 99(0)     & 93(2)     & 65(3)     & 0(0)      & 67(6)     & 41(5)     & 38(3)     & 0(0)      & 40(3)     & 0(0)      & 4(0)      & 0(0)      \\ 
             VAE & 98(1)     & 92(3)     & 64(3)     & 0(0)      & 67(6)     & 40(6)     & 39(3)     & 1(0)      & 43(4)     & 0(0)      & 3(0)      & 0(0)      \\ 
    \bottomrule 
\end{tabular} 

%% file: tables/CGO_2.tex
% ==================================================================== 
% This is an automatically generated tex file. 
% To update, please see: ./figures_and_tables/gen_table_CGO.py 
% ==================================================================== 
\begin{tabular}{lS[table-format=2(1)]S[table-format=2(1)]S[table-format=2(1)]S[table-format=2(1)]S[table-format=2(1)]S[table-format=2(1)]S[table-format=2(1)]S[table-format=2(1)]S[table-format=2(1)]S[table-format=2(1)]S[table-format=2(1)]S[table-format=2(1)]} 
    \toprule 
     & \multicolumn{2}{c}{position} & \multicolumn{2}{c}{hue} & \multicolumn{2}{c}{lightness} & \multicolumn{2}{c}{scale} & \multicolumn{2}{c}{shape} & \multicolumn{2}{c}{texture} \\ 
    \cmidrule(lr){2-3}\cmidrule(lr){4-5}\cmidrule(lr){6-7}\cmidrule(lr){8-9}\cmidrule(lr){10-11}\cmidrule(lr){12-13} 
                 & {FAAvg} & {FAMin} & {FAAvg} & {FAMin} & {FAAvg} & {FAMin} & {FAAvg} & {FAMin} & {FAAvg} & {FAMin} & {FAAvg} & {FAMin} \\ 
    \midrule 
            RN18 & 100(0)    & 100(0)    & 89(4)     & 53(16)    & 76(11)    & 51(14)    & 59(5)     & 21(6)     & 64(2)     & 4(2)      & 17(4)     & 0(0)      \\ 
            RN50 & 100(0)    & 100(0)    & 91(4)     & 59(17)    & 76(10)    & 56(14)    & 59(4)     & 18(7)     & 66(3)     & 13(5)     & 16(4)     & 0(0)      \\ 
         RN50-IN & 88(2)     & 67(6)     & 91(3)     & 80(4)     & 77(8)     & 63(12)    & 35(5)     & 15(4)     & 46(6)     & 18(5)     & 15(2)     & 2(1)      \\ 
         AlexNet & 97(2)     & 87(12)    & 87(2)     & 45(7)     & 75(9)     & 55(14)    & 49(4)     & 11(5)     & 55(2)     & 0(0)      & 7(2)      & 0(0)      \\ 
        DenseNet & 100(0)    & 98(2)     & 89(4)     & 56(14)    & 77(10)    & 58(15)    & 58(5)     & 23(4)     & 62(2)     & 8(3)      & 12(3)     & 0(0)      \\ 
             WRN & 100(0)    & 100(0)    & 91(4)     & 62(17)    & 78(11)    & 58(14)    & 59(4)     & 19(3)     & 67(3)     & 11(6)     & 17(4)     & 0(0)      \\ 
      ASR (RN18) & 99(1)     & 96(3)     & 39(6)     & 6(6)      & 55(7)     & 39(9)     & 62(2)     & 4(2)      & 65(3)     & 2(1)      & 14(2)     & 0(0)      \\ 
      Factor-VAE & 97(2)     & 89(6)     & 67(4)     & 1(0)      & 68(8)     & 43(10)    & 40(4)     & 1(0)      & 48(3)     & 0(0)      & 5(1)      & 0(0)      \\ 
             VAE & 99(1)     & 97(3)     & 69(4)     & 1(0)      & 68(7)     & 39(8)     & 41(4)     & 1(0)      & 51(4)     & 0(0)      & 4(0)      & 0(0)      \\ 
    \bottomrule 
\end{tabular} 

%% file: tables/CGO_3.tex
% ==================================================================== 
% This is an automatically generated tex file. 
% To update, please see: ./figures_and_tables/gen_table_CGO.py 
% ==================================================================== 
\begin{tabular}{lS[table-format=2(1)]S[table-format=2(1)]S[table-format=2(1)]S[table-format=2(1)]S[table-format=2(1)]S[table-format=2(1)]S[table-format=2(1)]S[table-format=2(1)]S[table-format=2(1)]S[table-format=2(1)]S[table-format=2(1)]S[table-format=2(1)]} 
    \toprule 
     & \multicolumn{2}{c}{position} & \multicolumn{2}{c}{hue} & \multicolumn{2}{c}{lightness} & \multicolumn{2}{c}{scale} & \multicolumn{2}{c}{shape} & \multicolumn{2}{c}{texture} \\ 
    \cmidrule(lr){2-3}\cmidrule(lr){4-5}\cmidrule(lr){6-7}\cmidrule(lr){8-9}\cmidrule(lr){10-11}\cmidrule(lr){12-13} 
                 & {FAAvg} & {FAMin} & {FAAvg} & {FAMin} & {FAAvg} & {FAMin} & {FAAvg} & {FAMin} & {FAAvg} & {FAMin} & {FAAvg} & {FAMin} \\ 
    \midrule 
            RN18 & 100(0)    & 100(0)    & 95(3)     & 81(9)     & 86(6)     & 65(13)    & 62(7)     & 22(3)     & 76(3)     & 28(7)     & 19(4)     & 1(0)      \\ 
            RN50 & 100(0)    & 100(0)    & 96(2)     & 85(8)     & 86(7)     & 66(17)    & 69(7)     & 29(9)     & 72(2)     & 8(3)      & 20(4)     & 1(0)      \\ 
         RN50-IN & 92(2)     & 74(7)     & 94(1)     & 81(5)     & 84(5)     & 70(10)    & 35(5)     & 12(3)     & 51(8)     & 23(10)    & 15(2)     & 3(1)      \\ 
         AlexNet & 97(2)     & 87(12)    & 86(2)     & 38(7)     & 79(7)     & 54(11)    & 56(7)     & 6(2)      & 62(4)     & 3(2)      & 10(2)     & 0(0)      \\ 
        DenseNet & 100(0)    & 100(0)    & 94(2)     & 80(6)     & 86(7)     & 71(13)    & 66(7)     & 23(4)     & 72(4)     & 29(11)    & 16(3)     & 1(0)      \\ 
             WRN & 100(0)    & 100(0)    & 97(2)     & 87(6)     & 87(6)     & 69(13)    & 74(6)     & 32(7)     & 82(3)     & 48(9)     & 24(6)     & 2(0)      \\ 
      ASR (RN18) & 100(0)    & 99(0)     & 45(1)     & 2(1)      & 62(8)     & 34(8)     & 67(3)     & 10(5)     & 69(3)     & 11(7)     & 14(1)     & 0(0)      \\ 
      Factor-VAE & 100(0)    & 100(0)    & 70(3)     & 1(1)      & 73(6)     & 36(7)     & 43(4)     & 1(1)      & 49(3)     & 0(0)      & 4(1)      & 0(0)      \\ 
             VAE & 100(0)    & 100(0)    & 72(3)     & 0(0)      & 72(7)     & 36(8)     & 44(5)     & 1(0)      & 53(4)     & 0(0)      & 3(1)      & 0(0)      \\ 
    \bottomrule 
\end{tabular} 

%% file: tables/FGO_95.tex
% ==================================================================== 
% This is an automatically generated tex file. 
% To update, please see: ./figures_and_tables/gen_table_FGO.py 
% ==================================================================== 
\begin{tabular}{lS[table-format=2(1)]S[table-format=2(1)]S[table-format=2(1)]S[table-format=2(1)]S[table-format=2(1)]S[table-format=2(1)]S[table-format=2(1)]S[table-format=2(1)]S[table-format=2(1)]S[table-format=2(1)]S[table-format=2(1)]S[table-format=2(1)]} 
    \toprule 
     & \multicolumn{2}{c}{position} & \multicolumn{2}{c}{hue} & \multicolumn{2}{c}{lightness} & \multicolumn{2}{c}{scale} & \multicolumn{2}{c}{shape} & \multicolumn{2}{c}{texture} \\ 
    \cmidrule(lr){2-3}\cmidrule(lr){4-5}\cmidrule(lr){6-7}\cmidrule(lr){8-9}\cmidrule(lr){10-11}\cmidrule(lr){12-13} 
                 & {FAAvg} & {FAMin} & {FAAvg} & {FAMin} & {FAAvg} & {FAMin} & {FAAvg} & {FAMin} & {FAAvg} & {FAMin} & {FAAvg} & {FAMin} \\ 
    \midrule 
            RN18 & 100(0)    & 100(0)    & 100(0)    & 100(0)    & 93(2)     & 85(8)     & 93(1)     & 89(1)     & 99(0)     & 98(0)     & 39(2)     & 35(3)     \\ 
            RN50 & 97(2)     & 87(12)    & 100(0)    & 100(0)    & 95(1)     & 93(2)     & 93(1)     & 90(1)     & 99(0)     & 98(0)     & 41(2)     & 36(1)     \\ 
         RN50-IN & 95(1)     & 91(2)     & 96(1)     & 94(1)     & 85(3)     & 78(4)     & 41(4)     & 28(4)     & 51(8)     & 37(9)     & 14(1)     & 4(1)      \\ 
         AlexNet & 100(0)    & 100(0)    & 97(2)     & 87(12)    & 82(4)     & 64(8)     & 82(3)     & 69(8)     & 94(3)     & 82(11)    & 14(2)     & 6(3)      \\ 
        DenseNet & 100(0)    & 100(0)    & 100(0)    & 100(0)    & 94(1)     & 92(2)     & 92(1)     & 88(1)     & 98(0)     & 97(1)     & 32(2)     & 26(2)     \\ 
             WRN & 100(0)    & 100(0)    & 100(0)    & 100(0)    & 95(1)     & 94(1)     & 95(1)     & 92(1)     & 98(0)     & 98(0)     & 47(3)     & 42(2)     \\ 
      ASR (RN18) & 99(0)     & 98(0)     & 34(3)     & 1(1)      & 53(4)     & 28(1)     & 69(2)     & 35(6)     & 70(2)     & 26(7)     & 10(1)     & 0(0)      \\ 
      Factor-VAE & 100(0)    & 100(0)    & 84(4)     & 61(9)     & 80(5)     & 69(7)     & 55(4)     & 40(4)     & 62(5)     & 44(7)     & 5(0)      & 0(0)      \\ 
             VAE & 100(0)    & 100(0)    & 86(3)     & 63(8)     & 82(4)     & 74(5)     & 57(4)     & 42(4)     & 68(5)     & 52(6)     & 4(0)      & 0(0)      \\ 
    \bottomrule 
\end{tabular} 

%% file: tables/FGO_90.tex
% ==================================================================== 
% This is an automatically generated tex file. 
% To update, please see: ./figures_and_tables/gen_table_FGO.py 
% ==================================================================== 
\begin{tabular}{lS[table-format=2(1)]S[table-format=2(1)]S[table-format=2(1)]S[table-format=2(1)]S[table-format=2(1)]S[table-format=2(1)]S[table-format=2(1)]S[table-format=2(1)]S[table-format=2(1)]S[table-format=2(1)]S[table-format=2(1)]S[table-format=2(1)]} 
    \toprule 
     & \multicolumn{2}{c}{position} & \multicolumn{2}{c}{hue} & \multicolumn{2}{c}{lightness} & \multicolumn{2}{c}{scale} & \multicolumn{2}{c}{shape} & \multicolumn{2}{c}{texture} \\ 
    \cmidrule(lr){2-3}\cmidrule(lr){4-5}\cmidrule(lr){6-7}\cmidrule(lr){8-9}\cmidrule(lr){10-11}\cmidrule(lr){12-13} 
                 & {FAAvg} & {FAMin} & {FAAvg} & {FAMin} & {FAAvg} & {FAMin} & {FAAvg} & {FAMin} & {FAAvg} & {FAMin} & {FAAvg} & {FAMin} \\ 
    \midrule 
            RN18 & 100(0)    & 100(0)    & 100(0)    & 100(0)    & 96(1)     & 95(1)     & 96(1)     & 95(1)     & 99(0)     & 99(0)     & 47(3)     & 41(4)     \\ 
            RN50 & 100(0)    & 100(0)    & 100(0)    & 100(0)    & 96(1)     & 95(1)     & 97(0)     & 95(1)     & 99(0)     & 99(0)     & 49(3)     & 41(2)     \\ 
         RN50-IN & 97(1)     & 95(1)     & 97(1)     & 96(1)     & 87(2)     & 83(3)     & 45(3)     & 35(4)     & 59(7)     & 48(9)     & 17(2)     & 8(1)      \\ 
         AlexNet & 97(2)     & 87(12)    & 95(3)     & 73(15)    & 90(3)     & 75(9)     & 89(1)     & 85(2)     & 95(3)     & 84(11)    & 28(2)     & 22(2)     \\ 
        DenseNet & 100(0)    & 100(0)    & 100(0)    & 100(0)    & 96(1)     & 95(1)     & 95(1)     & 94(1)     & 99(0)     & 98(1)     & 40(1)     & 36(1)     \\ 
             WRN & 100(0)    & 100(0)    & 100(0)    & 100(0)    & 97(1)     & 96(1)     & 97(0)     & 95(0)     & 99(0)     & 99(0)     & 54(5)     & 52(5)     \\ 
      ASR (RN18) & 99(0)     & 98(0)     & 45(4)     & 11(4)     & 53(3)     & 31(2)     & 78(1)     & 50(3)     & 74(4)     & 32(7)     & 10(0)     & 0(0)      \\ 
      Factor-VAE & 100(0)    & 100(0)    & 89(3)     & 72(9)     & 83(4)     & 76(6)     & 62(4)     & 51(5)     & 71(5)     & 60(5)     & 5(0)      & 0(0)      \\ 
             VAE & 100(0)    & 100(0)    & 91(2)     & 76(7)     & 86(3)     & 79(5)     & 64(4)     & 53(4)     & 78(5)     & 70(6)     & 4(0)      & 0(0)      \\ 
    \bottomrule 
\end{tabular} 

%% file: tables/FGO_80.tex
% ==================================================================== 
% This is an automatically generated tex file. 
% To update, please see: ./figures_and_tables/gen_table_FGO.py 
% ==================================================================== 
\begin{tabular}{lS[table-format=2(1)]S[table-format=2(1)]S[table-format=2(1)]S[table-format=2(1)]S[table-format=2(1)]S[table-format=2(1)]S[table-format=2(1)]S[table-format=2(1)]S[table-format=2(1)]S[table-format=2(1)]S[table-format=2(1)]S[table-format=2(1)]} 
    \toprule 
     & \multicolumn{2}{c}{position} & \multicolumn{2}{c}{hue} & \multicolumn{2}{c}{lightness} & \multicolumn{2}{c}{scale} & \multicolumn{2}{c}{shape} & \multicolumn{2}{c}{texture} \\ 
    \cmidrule(lr){2-3}\cmidrule(lr){4-5}\cmidrule(lr){6-7}\cmidrule(lr){8-9}\cmidrule(lr){10-11}\cmidrule(lr){12-13} 
                 & {FAAvg} & {FAMin} & {FAAvg} & {FAMin} & {FAAvg} & {FAMin} & {FAAvg} & {FAMin} & {FAAvg} & {FAMin} & {FAAvg} & {FAMin} \\ 
    \midrule 
            RN18 & 100(0)    & 100(0)    & 100(0)    & 100(0)    & 98(1)     & 97(1)     & 95(2)     & 84(11)    & 99(0)     & 99(0)     & 54(5)     & 51(5)     \\ 
            RN50 & 100(0)    & 100(0)    & 100(0)    & 100(0)    & 98(1)     & 97(1)     & 98(0)     & 98(0)     & 100(0)    & 99(0)     & 55(5)     & 52(5)     \\ 
         RN50-IN & 98(0)     & 98(1)     & 98(0)     & 97(1)     & 90(2)     & 88(2)     & 51(3)     & 44(3)     & 67(6)     & 60(7)     & 23(2)     & 15(2)     \\ 
         AlexNet & 97(2)     & 87(12)    & 100(0)    & 100(0)    & 90(4)     & 77(10)    & 91(3)     & 80(11)    & 93(5)     & 85(11)    & 38(2)     & 34(2)     \\ 
        DenseNet & 100(0)    & 100(0)    & 100(0)    & 100(0)    & 97(1)     & 96(1)     & 97(0)     & 97(0)     & 99(0)     & 99(0)     & 47(3)     & 43(3)     \\ 
             WRN & 100(0)    & 100(0)    & 100(0)    & 100(0)    & 98(0)     & 97(1)     & 98(0)     & 97(0)     & 99(0)     & 99(0)     & 61(5)     & 57(6)     \\ 
      ASR (RN18) & 100(0)    & 99(0)     & 59(4)     & 26(8)     & 60(3)     & 38(2)     & 84(2)     & 67(5)     & 82(3)     & 60(7)     & 16(1)     & 1(1)      \\ 
      Factor-VAE & 100(0)    & 100(0)    & 94(2)     & 84(5)     & 88(3)     & 82(5)     & 71(3)     & 63(3)     & 81(4)     & 75(5)     & 5(0)      & 0(0)      \\ 
             VAE & 100(0)    & 100(0)    & 96(1)     & 89(4)     & 89(3)     & 84(4)     & 72(3)     & 65(3)     & 85(3)     & 81(4)     & 5(0)      & 0(0)      \\ 
    \bottomrule 
\end{tabular} 